\documentclass[10pt,twocolumn,letterpaper]{article}
\pdfoutput=1
\usepackage{iccv}
\usepackage{times}
\usepackage{epsfig}
\usepackage{graphicx}
\usepackage{amsmath}
\usepackage{amssymb}
\usepackage{booktabs}
\usepackage{bbding}
\usepackage{pifont}
\usepackage{multirow}
\usepackage{subfigure}
\usepackage{array} 
\usepackage{multirow}
\usepackage{bm}
\usepackage{svg}
\usepackage{indentfirst}
\usepackage{cases}
\usepackage[ruled,vlined]{algorithm2e}
\usepackage{float}
\usepackage{stfloats}
\usepackage{mathrsfs}
\usepackage{pifont}

\usepackage[breaklinks=true,bookmarks=false]{hyperref}
\usepackage[capitalize]{cleveref}
\crefname{section}{Sec.}{Secs.}
\Crefname{section}{Section}{Sections}
\Crefname{table}{Table}{Tables}
\crefname{table}{Tab.}{Tabs.}
\Crefname{algorithm}{Algorithm}{Algorithms}
\crefname{algorithm}{Alg.}{Algs.}
\def\etal{{et al}\onedot}
\makeatother
\def\etalcite#1{\etal~\cite{#1}}

\iccvfinalcopy

\ificcvfinal\fi

\begin{document}

\title{NeRF-LOAM: Neural Implicit Representation for Large-Scale  \\ Incremental LiDAR Odometry and Mapping}

\author{Junyuan Deng$^1$
\quad
Xieyuanli Chen$^{2}$\thanks{corresponding authors}
\quad
Songpengcheng Xia$^1$
\quad
Zhen Sun$^1$
\\
Guoqing Liu$^1$
\quad
Wenxian Yu$^1$
\quad
Ling Pei$^{1,*}$
\\
$^1$Shanghai Jiao Tong University
$^2$College of Intelligence Science and Technology, NUDT
}

\maketitle

\ificcvfinal\thispagestyle{empty}\fi

\begin{abstract}
   Simultaneously odometry and mapping using LiDAR data is an important task for mobile systems to achieve full autonomy in large-scale environments. However, most existing LiDAR-based methods prioritize tracking quality over reconstruction quality. Although the recently developed neural radiance fields (NeRF) have shown promising advances in implicit reconstruction for indoor environments, the problem of simultaneous odometry and mapping for large-scale scenarios using incremental LiDAR data remains unexplored. 
   To bridge this gap, in this paper, we propose a novel NeRF-based LiDAR odometry and mapping approach, NeRF-LOAM, consisting of three modules neural odometry, neural mapping, and mesh reconstruction. All these modules utilize our proposed neural signed distance function, which separates LiDAR points into ground and non-ground points to reduce Z-axis drift, optimizes odometry and voxel embeddings concurrently, and in the end generates dense smooth mesh maps of the environment.
   Moreover, this joint optimization allows our NeRF-LOAM to be pre-trained free and exhibit strong generalization abilities when applied to different environments. 
   Extensive evaluations on three publicly available datasets demonstrate that our approach achieves state-of-the-art odometry and mapping performance, as well as a strong generalization in large-scale environments utilizing LiDAR data. Furthermore, we perform multiple ablation studies to validate the effectiveness of our network design.
   The implementation of our approach will be made available at \url{https://github.com/JunyuanDeng/NeRF-LOAM}.
       \vspace{-0.2cm}   
   \end{abstract}
   
      \begin{figure}[tb]
           \begin{center}
               \includegraphics[width=\linewidth]{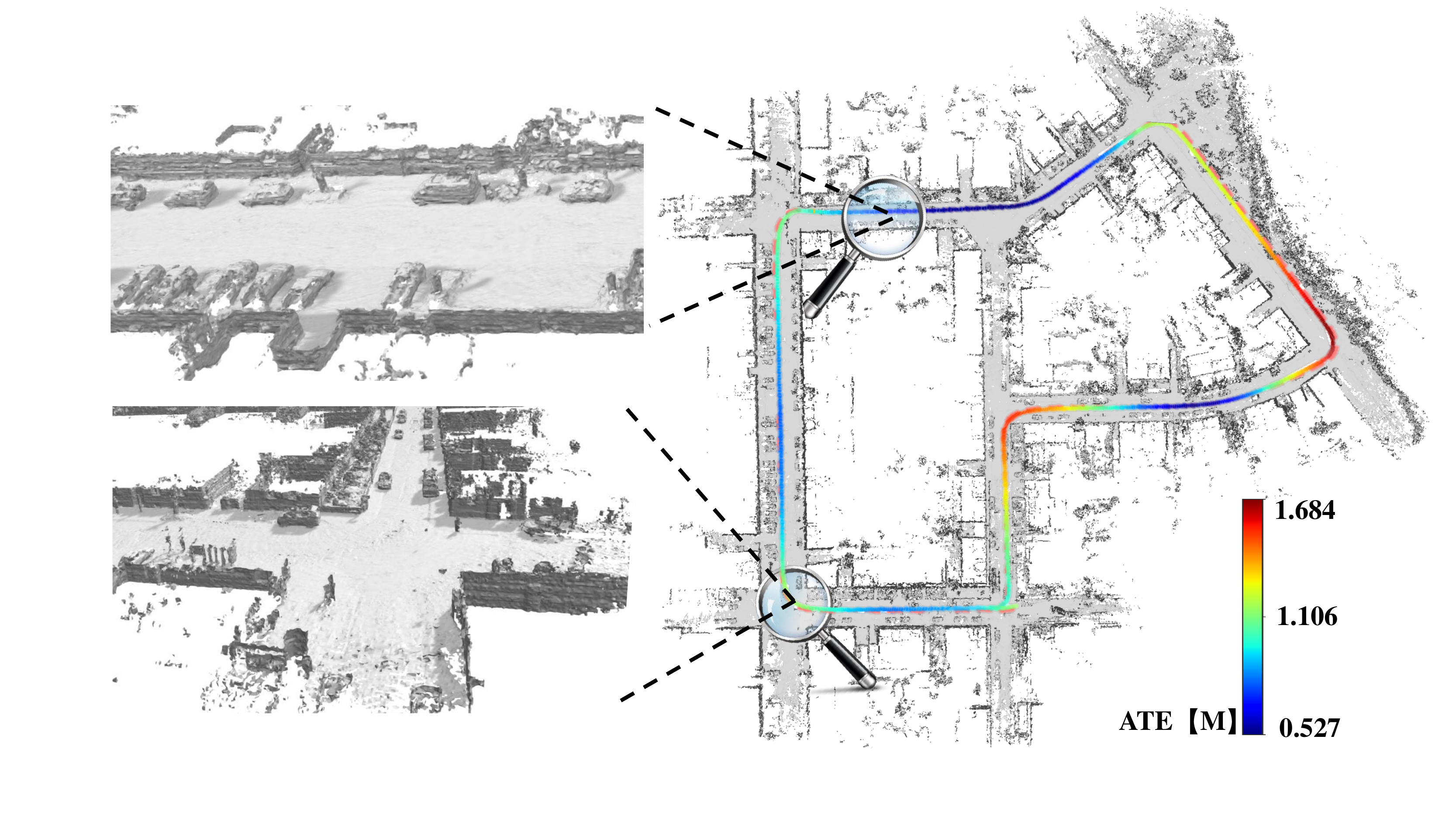}
           \end{center}
           \caption{Simultaneously odometry and dense mapping results on KITTI07. We present the reconstruction and the odometry result. The odometry results are colored by the absolute trajectory errors (ATE). Our proposed novel NeRF-LOAM accurately estimates the poses of a mobile system and reconstructs the dense mesh map of the outdoor large-scale environment.}
           \label{fig:motiv}
           \vspace{-0.4cm}
       \end{figure}

   \section{Introduction}
   
      Simultaneous odometry and mapping is an important component for autonomous mobile systems to achieve full autonomy in large-scale environments. It estimates the 6-degree-of-freedom poses of the vehicle and simultaneously builds a map of the environment, which are fundamental prerequisites for downstream tasks like path planning and collision avoidance. LiDAR sensors have been widely adopted for odometry and mapping due to their ability to provide precise range measurements and robustness to illumination changes. However, it can be argued that the current LiDAR odometry and mapping algorithms prioritize tracking quality over dense reconstruction quality, which may overlook the potential benefits of accurately capturing environmental geometry and generating high-fidelity reconstructions. 
       Despite the popularity of LiDAR-based incremental pose estimation~\cite{li2019cvpr,wang2019iros, nubert2021icra, wang2021cvpr}, research on high-level dense map reconstruction, especially deep-learning-based algorithms remains scarce.
       
       Recently, neural radiance fields (NeRF)~\cite{mildenhall2020eccv} has shown promising potentials in representing 3D scenes implicitly using a neural network and parallelly pose tracking methods~\cite{sucar2021iccv, zhu2022cvpr, yang2022ismar}. Although such representation can achieve seminal reconstruction with accurate poses, they concentrate on indoor pose tracking and scene representation with \mbox{RGB-D} sensors. The sparsity of LiDAR data and the lack of RGB information present significant challenges for applying previous algorithms to LiDAR data in outdoor environments. 
       Developing practical LiDAR-based algorithms to address these issues is currently a critical task.
      
      To this end, we propose a novel NeRF-based LiDAR odometry and mapping method, dubbed NeRF-LOAM. It employs sparse octree-based voxels combined with neural implicit embeddings, decoded into a continuous signed distance function (SDF) by a neural implicit decoder. The embeddings, decoder, and poses are optimized simultaneously by minimizing the SDF errors. NeRF-LOAM targets the outdoor driving environments and separates the LiDAR points into ground and non-ground points, and a precise SDF for ground points can be obtained with the help of normals. Such an operation depresses Z-axis drift and smooths our dense 3D map. To tackle the incremental odometry and mapping under the unknown large-scale outdoor environment, a dynamic voxel embedding generation strategy without any pre-allocation or time-consuming loop is designed. Finally, we use key-scans to not only jointly refine the pose and the map but also relieve the catastrophic forgetting or pre-training process. Extensive experiments were conducted on three publicly available datasets. The experimental results demonstrate that our method attains state-of-the-art odometry and mapping performance in outdoor large-scale environments using LiDAR data. 
   
       To sum up, the contributions of our work are threefold:
       \begin{enumerate}
       \vspace{-0.2cm}
           \item To the best of our knowledge, our NeRF-LOAM is the first neural implicit odometry and mapping method for large-scale environments using LiDAR data.
       \vspace{-0.2cm} 
           \item We propose a novel neural SDF module combined with dynamic generation and key-scans refine strategy, which realizes a fast allocation of voxel embeddings in the octree and a high-fidelity 3D representation.
       \vspace{-0.2cm} 
           \item Based on the proposed online joint optimization, our method is pre-training free and generalizes well in different environments.
       \end{enumerate}

       \section{Related Work}
       Odometry and mapping in outdoor large-scale environments using LiDAR data has been investigated for decades. One of the primary methods is the iterative closest point (ICP) algorithm~\cite{besl1992pami, Rusinkiewicz20013di}, which directly aligns consecutive point clouds together and calculates the relative transformation between pairs of LiDAR scans. Tackling the sparsity of LiDAR data, Zhang and Singh~\cite{zhang2014rss} use point-to-edge and point-to-plane distance to optimize the ICP error and achieve accurate odometry estimates. However, these types of algorithms mainly focus on odometry estimation, while the reconstructed map is coarse. The successive research~\cite{besl1992pami, segal2009rss,behley2018rss, chen2019iros} also explores the scene geometry to get more accurate odometry results without considering the quality of the reconstruction map. Meanwhile, learning-based methods on LiDAR odometry~\cite{li2019cvpr,wang2019iros, nubert2021icra, wang2021cvpr, chen2020iros} attract much attention. These methods employ a network to learn features from points or projected 2D images. However, they often require large data for training and cannot generalize well to other environments.
       
       To represent the 3D scene, there are many techniques such as surfels~\cite{pfister2000siggraph}, occupancy grids~\cite{1989Using}, triangle meshes~\cite{lorensen1987siggraph,chen2021icra}, and polynomial representations~\cite{kolluri2008talg}. 
       Traditionally, Poisson surface reconstruction~\cite{kazhdan2006eg,kazhdan2013acmgraphics} provides geometrically accurate reconstruction. Newcombe~\etalcite{izadi2011acmsuist} popularizes the concept of truncated signed distance function (TSDF) and volumetric integration methods to reconstruct triangle meshes~\cite{klingensmith2015rss, vizzo2022sensors}. Behley and Stachniss~\cite{behley2018rss} use surfels to realize the reconstruction of 3D range sensors. For learning-based reconstruction, they usually focus on the small objects~\cite{ma2022cvpr} or reconstruct directly from the point clouds~\cite{wiesmann2021ral} as a map database. The dense reconstruction from 3D incremental LiDAR data still remains to explore. 
       
       Compared to the existing 3D representations, the success of neural implicit representation~\cite{azinovic2022cvpr, liu2020nips, mildenhall2020eccv, wang2021arXiv, zhong2023icra} for novel view synthesis attach great attention, and many research investigates the possibility to use this concept realizing simultaneous localization and mapping (SLAM)~\cite{wangzi2021arXiv,yen2020iros,ortiz2022arxiv, sucar2021iccv, zhu2022cvpr, yang2022ismar}. 
       These neural SLAM use multilayer perceptrons (MLPs) to represent the entire scene and achieve seminal results. 
       Extensive related works have been done such as the training and inference speed~\cite{liu2020nips, Lindell2021cvpr}, sparse training view~\cite{yu2020cvpr,Chen2021cvpr} and scene composition\cite{kaizhang2020arXiv, xie20213dv}.
       However, they are mainly designed to process the image~\cite{ mildenhall2019tog, mildenhall2020eccv} or RGBD inputs~\cite{dai2017cvpr,change20173dv} and are employed indoors. Extending them to LiDAR-based outdoor environments is hard to achieve because of the model limitation of simple MLPs and the sparsity character of LiDAR data. 
       Although~\cite{yang2022ismar, zhong2023icra} adopt an octree-based sparse grid with voxel embeddings and can be applied in larger areas, the pre-allocated embeddings or time-consuming loop to search the voxels is not available in outdoor for both odometry and mapping.
   
       Unlike the above-mentioned methods, we propose a novel neural implicit odometry and mapping method for incremental LiDAR inputs under large-scale environments to obtain both dense 3D representation and accurate poses. We adopt voxel embeddings with an MLP decoder to represent the local geometry instead of the entire scene, which generalizes well in most environments. We also design a dynamic voxel embedding generation strategy to reduce processing time significantly as well as a key-scans refine strategy to improve the reconstruction quality.

   \section{Our Neural SDF}
   \label{subsec:neusdf}
       
\begin{figure}[t]
           \begin{center}
               \includegraphics[width=\linewidth]{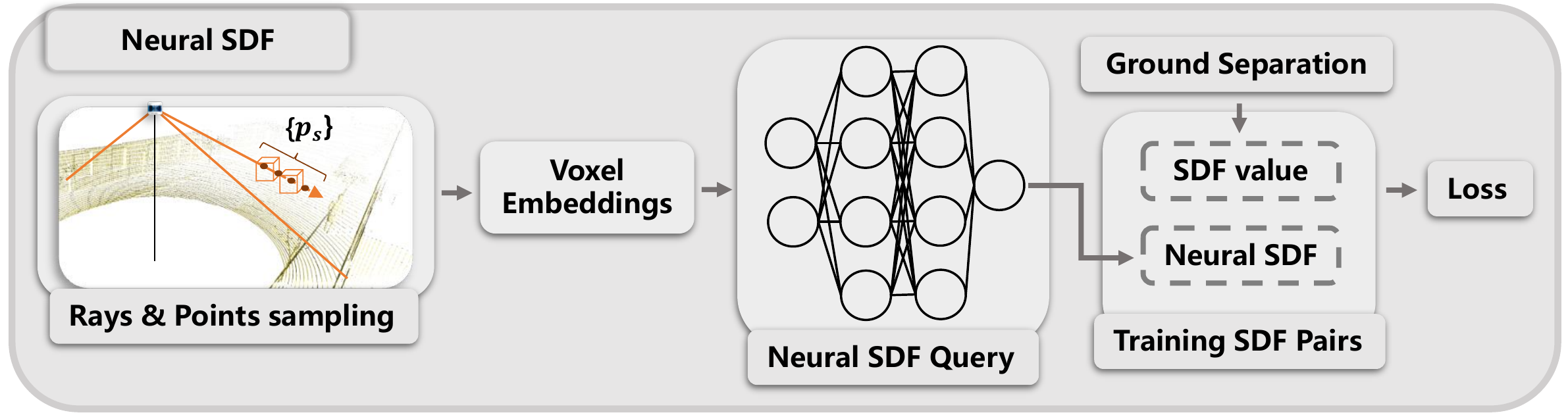}
               \caption{The modified neural SDF. After the rays and points sampling, the voxel embeddings are fed to a network to query the neural SDF after ground separation.}
               \label{fig:neuralsdf}
           \end{center}
           \vspace{-0.5cm}
       \end{figure}

       Before delving into the details of our NeRF-LOAM network, we first introduce a novel neural SDF module shown in~\cref{fig:neuralsdf}, which plays a crucial role in all of our processes, including optimizing the poses, maps, and networks. 
   
       To realize the neural representation of large-scale outdoor incremental, the octree~\cite{meagher1980techreport, vespa2018ral, takikawa2021cvpr} structure is often adopted to recursively divide the scene into leaf nodes with basic scene units voxels. These axis-aligned voxels attach an $N_e$-dimension embedding at each vertex and share with neighbor voxels. The SDF values can be inferred from the embeddings through a neural network $F_\theta$.
       Different from existing methods~\cite{yang2022ismar, zhong2023icra}, we treat the environments differently when optimizing the SDF values, e.g., ground and non-ground, and propose a novel loss function to realize more suitable neural SDF for LiDAR data in outdoor large-scale environments.
   
       {\bf Rays and points sampling.} 
       The first step in all of our processes is based on effective sampling. Instead of randomly selecting samples across the space or around the points, we first select rays that intersect with the currently allocated voxel and then select the points along the intersection part of the ray and voxels. Note that we set a hit number threshold of voxels $M_n$ to avoid the influence of the unseen surface. Since the LiDAR rays are transformed by the scan pose $\bm{T}_i$, each ray contains the pose information of the scan. This sampling strategy allows us to optimize the pose and voxel embeddings simultaneously.
   
       {\bf Neural SDF value.} 
       For most visual-based NeRFs~\cite{azinovic2022cvpr, zhu2022cvpr}, the scalar value like weight or color are obtained by differentiable rendering along the ray. Since the SDF is a direct method to represent the scene, the rendering method is unsuitable for LiDAR data in outdoor environments~\cite{zhong2023icra}. The neural SDF filed $\Psi: \mathbb{R} ^3 \to \mathbb{R}$ can be represented by~\cref{eq:neusdf}: each sampled point can be regressed via the trilinear interpolation of voxel embeddings:
      \begin{equation}
         \Psi(\bm{p}_s) = F_\theta(\text{TriInpo}(\bm{p}_s, \bm{e}_1, ..., \bm{e}_n)),
      \label{eq:neusdf}
      \end{equation}
       where $\bm{p_s}=\bm{T}_i\bm{p}_f$ is the transformed sampled points by current scan pose $\bm{T}_i$ from the original point $\bm{p}_f$ in LiDAR coordinate, $\text{TriInpo}(\bm{p}_s, \bm{e}_1, ..., \bm{e}_n)$ represents the trilinear interpolation of the sampled point $\bm{p_s}$ surrounded by $n$ neighbor voxel embeddings, and $F_\theta$ is the neural implicit network with parameter $\theta$. 
       Since all processes involved are differentiable, we can optimize the scan pose, voxel embeddings, and network parameters jointly through back projection. Because the voxel embeddings primarily store geometric information, our network does not require pre-training and can adjust online to different environments.
   
       \begin{figure}[tb]
           \begin{center}
               \includegraphics[width=\linewidth]{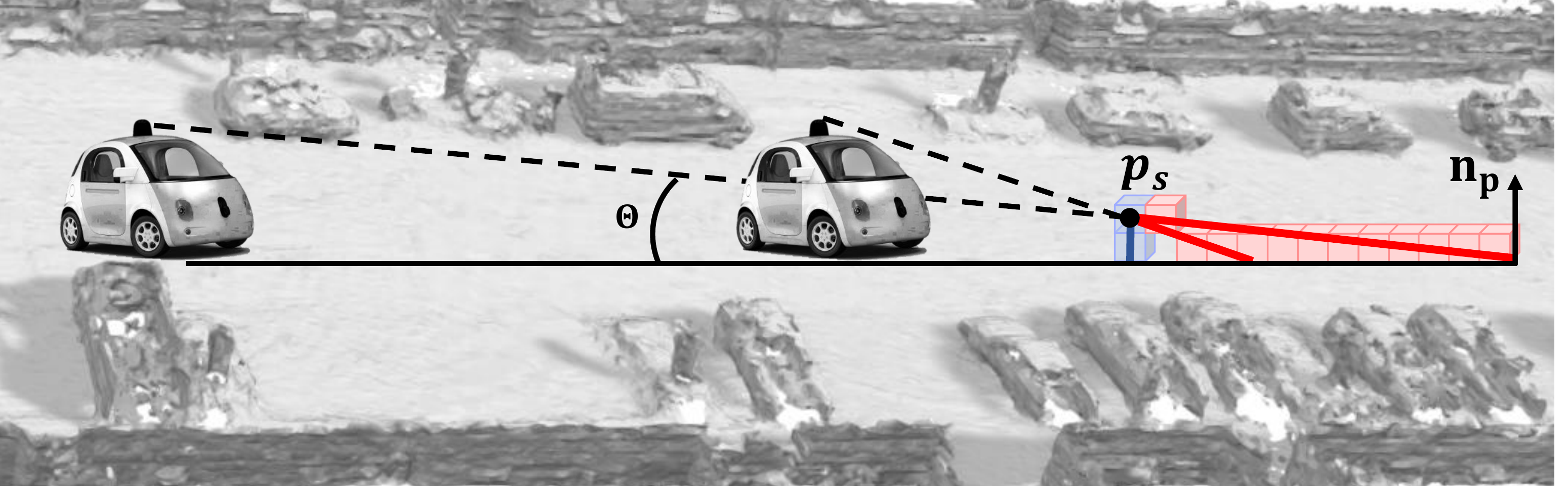}
           \end{center}
           \caption{The geometric information SDF value at point $\bm{p}_s$ should stay invariant w.r.t the view (blue line). While the approximated SDF is significantly different with view change (red line). The alignment of SDF forces the car to shift along the ray. }
           \label{fig:groundpoint}
           \vspace{-0.4cm}
       \end{figure}
   
       {\bf Training SDF pairs.} 
       The LiDAR sensors provide highly accurate range measurements, which allow us to compute the signed distance from the sampled points to the endpoints along the ray. This signed distance is often called the SDF value in many SLAM or mapping approaches~\cite{vizzo2022sensors, klingensmith2015rss}. This approximation is generally acceptable for simple mapping or indoor SLAM tasks while leading to sub-optimal results when applied to outdoor SLAM as shown in~\cref{fig:groundpoint}. It illustrates the issue with the SDF approximation when used with a far LiDAR point. The blue line represents the true SDF value, while the red line is the SDF approximation. The difference between the two distances can be significant when the angle $\theta$ is close to $0^{\circ}$. 
       This can decrease odometry quality due to the inaccurate SDF value. This problem is even more significant in the Z-axis, as there are fewer points in LiDAR scan to constrain Z-drift. While obtaining the normals of all LiDAR points can be challenging, the ``smooth" ground allows access to the rectified SDF value.
       
       Therefore, we propose to first separate LiDAR points into ground points $\mathcal{G}$ and non-ground points $\mathcal{G}^\mathcal{C}$. The SDF field $\Phi :\mathbb{R} ^3 \to \mathbb{R}$ can then be represented as:
       \begin{equation}
           \Phi(p_s)=\begin{cases}
               (\bf{p_s} - \bf{p}) \bf{n_p} &\text{if\space} p\in\mathcal{G}\\
              \left \| \bf{p_s} - \bf{p} \right \| & \text{else}
              \end{cases},
              \label{eq:sdf}
       \end{equation}
       where $\bf{p_s}$ is the sampled point and $\bf{p}$ is the LiDAR point alone the ray, $\bf{n_p}$ is the normal of point $\bf{p}$.
   
       {\bf Optimization.} 
       We train the network using the weighted sum of three different losses. The first free space loss forces the neural SDF of points between LiDAR and the positive truncation region $\mathcal{P}_{f} $ to be truncation distance $Tr$: 
      \begin{equation}
         \mathcal{L}_{f} = \frac{1}{|P_{f}|}\sum_{i=0}^{|P_{f}|}( \Psi(\bm{p}_i) - Tr)^2
      \end{equation}
       The negative truncation region is beyond our consideration following the suggestion of ~\cite{azinovic2022cvpr} to avoid surface intersection ambiguities~\cite{wang2021arXiv}. 
       This loss plays an important role in removing dynamic objects. Secondly, we define an SDF loss of points within the truncation region $\mathcal{P}_{s}$ to supervise the SDF estimates:
       \begin{equation}
           \mathcal{L}_{s} = \frac{1}{|P_{s}|}\sum_{i=0}^{|P_{s}|}( \Psi(\bm{p}_i) - \Phi(\bm{p}_i))^2.
       \end{equation}
       
       Different to~\cite{ortiz2022arxiv,zhong2023icra} using a sigmoid function to increase the credits around the LiDAR points, we treat the points equally in this region for the reason that these points are all important for odometry. Finally, because the SDF values are differentiable and equal to one within the truncation area, we add an Eikonal loss:
       \begin{equation}
           \mathcal{L}_{e} = \frac{1}{|P_{s}|}\sum_{i=0}^{|P_{s}|}( \frac{\partial  \Psi(\bm{p}_i)}{\partial \bm{p}_i} -1 )^2 .
       \end{equation}

   \section{NeRF-LOAM Framework}
   
   \begin{figure}[t]
       \begin{center}
           \hspace{-0.15cm}\includegraphics[width=1.02\linewidth]{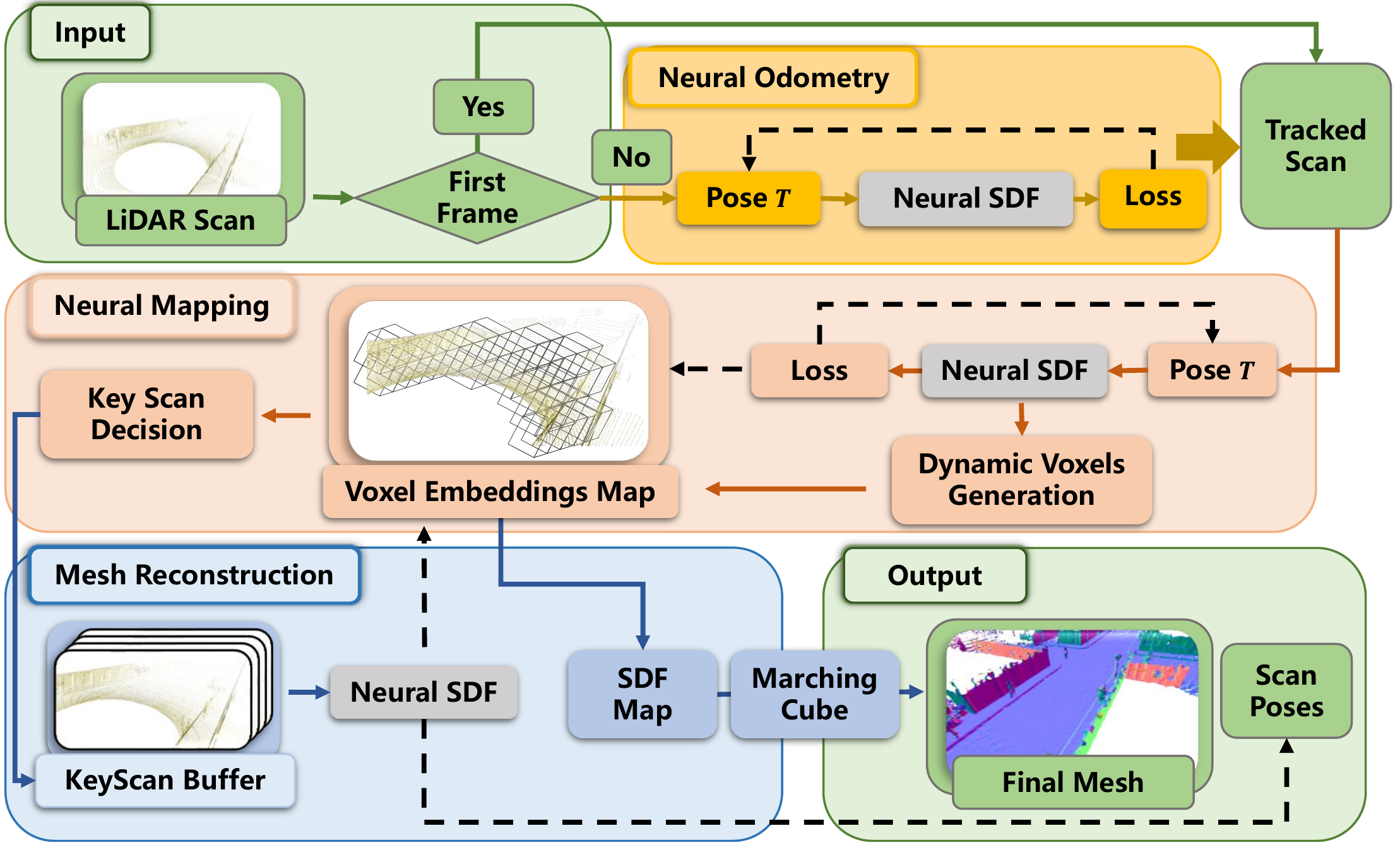}
           \caption{Our NeRFLOAM Overview. The dashed line represents the back projection. Given a LiDAR stream, our approach outputs the poses of each scan and a reconstructed mesh map of the environment with three modules: 1) neural odometry takes the pre-processed scan and optimizes the pose via back projecting the queried neural SDF; 2) neural mapping jointly optimizes the voxel embeddings map and pose while selecting the key-scans; 3) key-scans refined map returns SDF value and the final mesh is reconstructed by marching cube.}
           \label{fig:overview}
       \end{center}
           \vspace{-0.6cm}
   \end{figure}

   \subsection{Overview}
      The architecture of our framework is illustrated in~\cref{fig:overview}. Our method takes an incremental LiDAR stream as input and outputs a 3D reconstructed mesh with poses of each LiDAR scan through three modules: neural odometry, neural mapping, and mesh reconstruction. The first two parts run parallel as frontend and backend, while the third runs separately to obtain a global mesh map and refined scan poses.
      
       Given the incoming LiDAR scan $\bm{P}_t = \left \{\bm{p}_i \in \mathbb{R} ^{3}\right \}^{N}_{i=1}$, the neural odometry estimates a 6-DoF Pose $\bm{T}\in SE(3)$ for that scan by minimizing the SDF error from a fixed implicit network $F_\theta$ (see~\cref{subsec:no}). The tracked scan is then fed to neural mapping, which utilizes the tracked pose $\bm{T}$ to transform the point cloud into the world coordinate system (see~\cref{subsec:neuralmap}). The implicit map representation and pose are then jointly optimized. During mapping, we add a scan into the key-scan buffer after a certain distance or when the vehicle reaches a new area. This key-scan buffer maintains the long-map consistency but also enhances the mapping quality.
      Finally, the key scans are utilized to refine both the odometry and map results (see~\cref{subsec:meshrecon}). The 3D mesh is reconstructed by the marching cube method~\cite{lorensen1987siggraph} based on the SDF values predicted by our network. More details of each component are provided in the following sections.

   \subsection{Neural Odometry}
   \label{subsec:no}
       For every incoming LiDAR scan $\bm{P}_t$, we randomly select $N$ rays and transform them into the world coordinate system. A set of points are sampled along the ray as described in~\cref{subsec:neusdf}. The pose and voxel embeddings are optimized by decreasing the loss. 
       
      For our neural odometry, the parameter which needs to be optimized is the 6-DoF pose $\bm{T}$ in $SE(3)$ space. All updates of the pose $\bm{\xi} \in \mathfrak{se}(3)$ is performed in tangent space of $SE(3)$. The Lie-algebra representation enables us to update the pose by a gradient descent method. We randomly select $N$ rays and transform them into the world coordinate system. Note that we use a constant move model to initialize our pose. This model can relieve our learning burden. We sample the points, compute the loss and optimize the pose via back-projection, as mentioned in~\cref{subsec:neusdf}. Here, the voxel embeddings and the network are obtained after the neural mapping process of the last tracked scan. 
   
       To tackle the problem of catastrophic forgetting when performing online incremental odometry, we freeze the network parameters after $K$ scans which does not decline our result because local geometry is mainly stored in voxels. The voxel embeddings and poses of the first $K$ scans will be refined later by key scan refinement, detailed in~\cref{subsec:meshrecon}.

   \subsection{Neural Mapping}
   \label{subsec:neuralmap}
   
      {\bf Dynamic voxel embeddings generation.}
       For neural mapping, we employ an octree-based approach to partition the scene. Following the odometry process, the estimated pose enables us to convert all points of the current scan into the world coordinate system. Subsequently, any points not in existing voxels are assigned to newly generated ones. These voxels are added to the octree along with their corresponding voxel embeddings. To quickly locate the desired embeddings, we encode the 3D voxel coordinates into a unique scalar value, namely the Morton code~\cite{vespa2018ral}. Although utilizing the code, the pre-allocate embeddings ~\cite{takikawa2021cvpr,yang2022ismar} or time-consuming on by one search in hash table~\cite{zhong2023icra} is not suitable for our task, especially when it needs to retrieve hundreds of thousands of embeddings from a hash table containing millions of entries.
   
       Inspired by the concept of a look-up table, we devise an efficient and scalable method for generating voxel embeddings dynamically, as outlined in~\cref{alg:algorithm1}. The lookup table is extended with the maximum Morton code to store the access information of voxels. The unvisited voxels will be assigned initialized embeddings and added to the embedding list while updating the look-up table by the current embedding number, eliminating time-consuming loop queries.

       {\bf Joint optimization of the map and pose.} 
       Similar to neural odometry, we sample the rays and points to calculate the loss. Here we mainly optimize the voxel embeddings while fine-tuning the poses.
   
      \begin{algorithm}[t]
         \caption{Dynamic Embeddings Generation.}
         \LinesNumbered
         \label{alg:algorithm1}
         \KwIn{Look-up list $\mathcal{L}$; Incoming voxels IDs (i.e., Morton code) $\mathcal{I}_v$; Embedding list $\mathcal{L}_e$;}
         \KwOut{Embedding list $\mathcal{L}_e$ with new embeddings; Updated look-up list $\mathcal{L}$.}  
         \BlankLine
         $m$ $\leftarrow$max($\mathcal{I}$), maximum index.
         
         $l \leftarrow $ len($\mathcal{L}$), length of the look-up list.
   
         $s \leftarrow $ len($\mathcal{L}_e$), length of the embeddings list.
   
         \If{$m > l$}
            {Extend the length of the look-up list to m, initialized with value -1.}
         $\mathcal{I}_e \leftarrow \mathcal{L}[\mathcal{I}_v]$, look the embeddings IDs.
   
         $\mathcal{I}_v \leftarrow \{ \mathcal{I}_v[i]$ \space $\vert $\space$\mathcal{I}_e[i]=-1 \}$, unvisited voxels IDs.

         $l_v \leftarrow$ len($\mathcal{I}_v$), length of unvisited voxels.
   
         $ \mathcal{L}_e^{\prime}\leftarrow[\bf{e}_i\vert i \in\{1,..., l_v\}]$, new embeddings.
   
         $\mathcal{L}_e \leftarrow \mathcal{L}_e + \mathcal{L}_e^{\prime}$, final embedding list.

         $\mathcal{L}[\mathcal{I}_v] \leftarrow [s+1, ..., s+l+v]$, update the look-up list.
   
      \end{algorithm}

   \subsection{Mesh Reconstruction}
   \label{subsec:meshrecon}
       {\bf Key-scans selection and refinement.}
       We maintain a key-scan buffer to relieve the catastrophic forgetting of the first $K$ scans as well as improve the mapping quality. A key scan is added to the buffer if the number of newly added voxels $N_v$ exceeds a threshold of $N_t$ or the distance between the current scan and the last key-scan $d_f$ is sufficiently large.  
       The map and poses are in the end refined with all the key-scans in the buffer.
       This simple strategy is effective, as demonstrated in the mapping results in~\cref{subsect:abla}. Additionally, to improve the efficiency of the refinement process, only rays or LiDAR points within a truncation distance $d_t$ based on the point density are included.
       
       {\bf Final mesh and poses.} 
       After the key-scans refine, the map and the poses are well-trained and ready to output final results. Our modified SDF is continuous, so we can theoretically infer SDF values at an arbitrary position. We query the SDF values with the same fixed size (i.e., voxel size), and the final mesh is obtained via marching cube~\cite{lorensen1987siggraph}.

   \section{Experiments}
       \subsection{Experimental Setup}
       {\bf Datasets.} 
       We evaluate our method and compare it with state-of-the-art (SOTA) methods using three publicly available outdoor LiDAR datasets, including MaiCity~\cite{vizzo2021icra}, Newer College~\cite{ramezani2020iros}, and KITTI odometry~\cite{geiger2012cvpr} datasets. MaiCity~\cite{vizzo2021icra} contains 64-beam noise-free synthetic LiDAR scans in urban environments, and the ground truth map is provided. Newer College~\cite{ramezani2020iros} contains a hand-carried LiDAR sequence collected at Oxford University with motion distortion. To make it more challenging and the scans more distinctive, we take one out of every five. We compare our odometry and mapping results with provided ground truth trajectories and mesh maps by these two datasets. KITTI odometry~\cite{geiger2012cvpr} does not provide ground truth maps, so we present our odometry accuracy hereby qualitative mapping results.
      
       {\bf Evaluation metric.} 
       We evaluate both the odometry and mapping performance of our method. For odometry accuracy, we present the root-mean-square error (RMSE) of absolute trajectory errors (ATEs) by $SE(3)$ alignment. And for mapping accuracy, we use the commonly used reconstruction metrics adopted in most reconstruction method~\cite{mescheder2019cvpr, vizzo2021icra, zhong2023icra}, i.e., accuracy, completion, Chamfer-L1 distance, and F-score, obtained by comparing the resulting mesh with ground truth. 
      
       {\bf Implemental details.} 
       The whole process shared network is an MLP consisting of 2 FC layers, and each layer has 256 hidden units. The length of our voxel embeddings is 16 with a voxel size 0.2\,m. For sampling, we set the step size ratio to 0.2 for odometry and 0.5 for mapping and the truncation distance $Tr = 0.3$\,m. To distinct the ground from the LiDAR points, we use the seminal work of~\cite{lim2021ral}. More studies on our hyperparameter selection are presented in~\cref{subsect:abla} and supplementary materials (see Sec. D).

   \subsection{Simultaneously Odometry\,\&\,Mapping Results}
      
      \begin{table*}
         \begin{center}
            \resizebox{\linewidth}{!}{
            \begin{tabular}{lc|cccc|cccc}
               \toprule
               \multirow{2}{*}{Method}& \multirow{2}{*}{Pose} & \multicolumn{4}{c|}{MaiCity}& \multicolumn{4}{c}{Newer College}\\

               &&Map. Acc. $\downarrow$&Map. Comp. $\downarrow$&C-l1. $\downarrow$& F-score (10cm) $\uparrow $&Map. Acc. $\downarrow$&Map. Comp. $\downarrow$&C-l1. $\downarrow$& F-score (20cm) $\uparrow $\\

               \toprule
               SHINE~\cite{zhong2023icra}&\multirow{3}{*}{KissICP~\cite{vizzo2023ral}} & 5.75
               & 38.45 &22.10& 67.00 & 14.87
               & \textbf{20.02} &\textbf{17.45}& 68.85\\
               Vdbfusion~\cite{vizzo2022sensors}&&4.95&46.79&25.87& 68.15&\textbf{14.03}&25.46&19.75&\textbf{69.50}\\
               Ours     &&\textbf{4.16}&\textbf{37.20}&\textbf{20.67}& \textbf{73.31}&14.31&24.39&19.35& 68.70\\

               \toprule
               Puma~\cite{vizzo2021icra}&\multirow{2}{*}{Odometry} & 7.89 &\textbf{9.14}& 8.51& 68.04& 15.30&71.91& 43.60& 57.27\\
               Ours&&\textbf{5.69}&11.23&\textbf{8.46}& \textbf{77.26} &\textbf{12.89}&\textbf{22.21}&\textbf{17.55}& \textbf{74.37}\\
               \bottomrule
            \end{tabular}
            }
         \end{center}
           \vspace{-0.2cm}
         \caption{Simultaneously odometry\,\&\,mapping results of different methods on MaiCity~\cite{vizzo2021icra} and Newer College ~\cite{ramezani2020iros} datasets in terms of map accuracy, completion and Chamfer-L1 distance and F-scores.}
         \label{tab:odome_mapping}
           \vspace{-0.4cm}
      \end{table*}

      \begin{figure*}[t]
         \subfigure[Ours with KissICP]{
            \centering
            \includegraphics[width=0.15\textwidth]{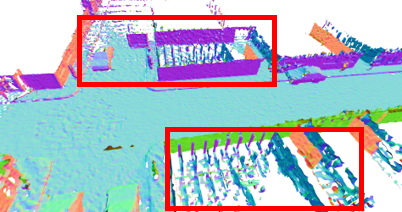}
         }
         \subfigure[Vdb with KissICP]{
            \centering
            \includegraphics[width=0.15\textwidth]{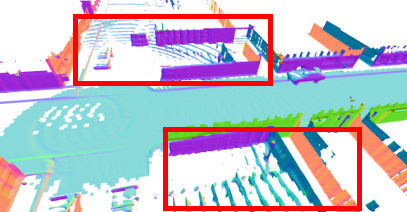}
          }
          \subfigure[Shine With KissICP]{
            \centering
            \includegraphics[width=0.15\textwidth]{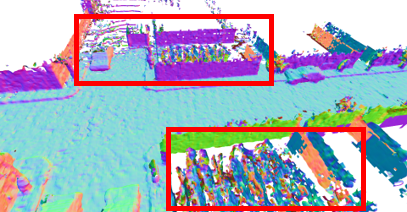}
         }
          \subfigure[Ours with KissICP]{
             \centering
             \includegraphics[width=0.15\textwidth]{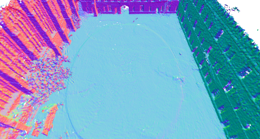}
          }
          \subfigure[Vdb with KissICP]{
            \centering
            \includegraphics[width=0.15\textwidth]{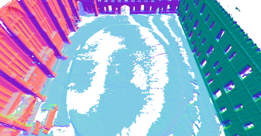}
         }
         \subfigure[Shine With KissICP]{
            \centering
            \includegraphics[width=0.15\textwidth]{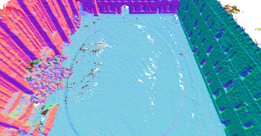}
         }
           \vspace{-0.1cm}
         \caption{The Odometrey mapping results for different methods. The first three are on MaiCity~\cite{vizzo2021icra} while the last three are on Newer College~\cite{ramezani2020iros}. The artifacts are highlighted in Red boxes.}
         \label{fig:kissicp_quali}
           \vspace{-0.2cm}
      \end{figure*}

      \begin{figure*}[t]
         \subfigure[Ours with GT pose]{
            \centering
            \includegraphics[width=0.185\textwidth]{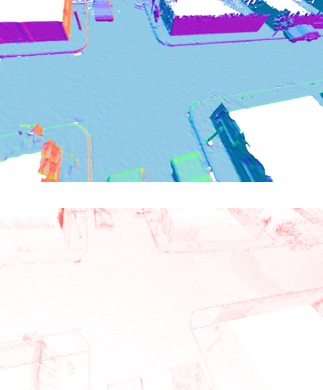}
         }
         \subfigure[Vdb with GT pose]{
            \centering
            \includegraphics[width=0.185\textwidth]{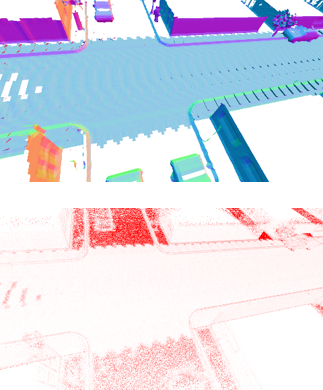}
          }
          \subfigure[Shine With GT pose]{
            \centering
            \includegraphics[width=0.185\textwidth]{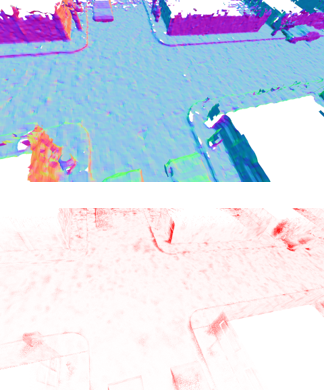}
         }
         \subfigure[Our odometry mapping]{
            \label{fig:odo_map_quli_d}
             \centering
             \includegraphics[width=0.185\textwidth]{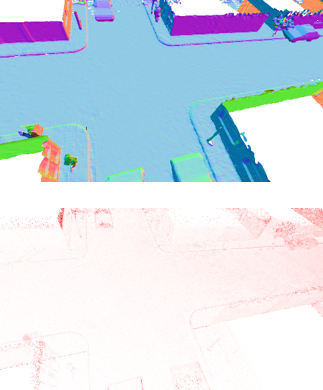}
          }
          \subfigure[Puma odometry mapping]{
            \label{fig:odo_map_quli_e}
            \centering
            \includegraphics[width=0.183\textwidth]{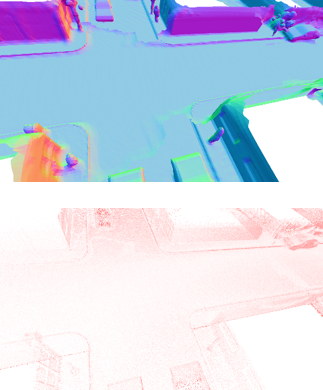}
         }
         \caption{The mapping result with ground truth pose or odometry results on the MaiCity~\cite{vizzo2021icra} dataset are shown in the first row. The second  row presents the error maps with ground truth mesh as a reference, where the redder points mean larger error up to 25cm.}
         \label{fig:odo_map_quli}
           \vspace{-0.1cm}
      \end{figure*}
   
        The first experiment shows the simultaneous odometry and dense mapping results of our method compared with existing SOTA methods. 
        For example, Poisson surface reconstruction SLAM method Puma~\cite{vizzo2021icra}, a TSDF fusion-based approach Vdbfusion~\cite{vizzo2022sensors}, and an implicit neural network-based map representation SHINE-Mapping~\cite{zhong2023icra}. 
        Since both Vdbfusion and SHINE-Mapping only focus on dense mapping, we combine them with the current SOTA odometry method KissICP~\cite{vizzo2023ral}. For fair comparison, we also show the results of our methods using KissICP poses.
        The results of all baseline methods are produced using their official open-source code with the same voxel size. 
        
        \cref{tab:odome_mapping} shows our odometry mapping results on the MaiCity\cite{vizzo2021icra} and Newer College\cite{ramezani2020iros} datasets. As can be seen, our mapping process combined with KissICP outperforms all baselines on the MaiCity dataset and has comparable quality in the Newer College dataset.
        The corresponding qualitative results are demonstrated in~\cref{fig:kissicp_quali}. 
        In the case of the MaiCity dataset, KissICP produces false pose estimates in the initial scans, which will lead to entangled mapping if there are no specific processes to remove these artifacts. Vdbfusion provides space carving to address this problem. However, it removes both the artifacts and important objects such as roads, trees, and cars. 
        Shine-Mapping offers some improvement by removing certain artifacts. Our proposed method outperforms both of these techniques by effectively removing the majority of the artifacts and producing a smoother mapping result.
        Similar benefits can be observed in the Newer College dataset, where Vdbfusion removes the trajectory caused by a person holding a device, resulting in an incomplete map.  
        
\begin{figure*}[t]
         \subfigure[Ours with GT pose]{
            \centering
            \includegraphics[width=0.18\textwidth]{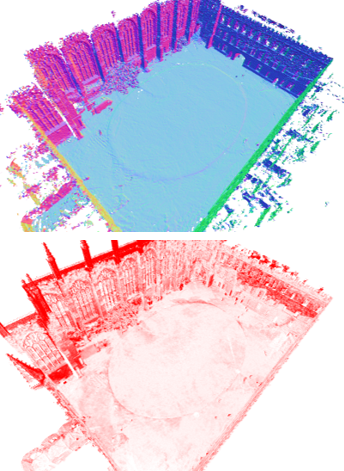}
         }
         \subfigure[Vdb with GT pose]{
            \centering
            \includegraphics[width=0.18\textwidth]{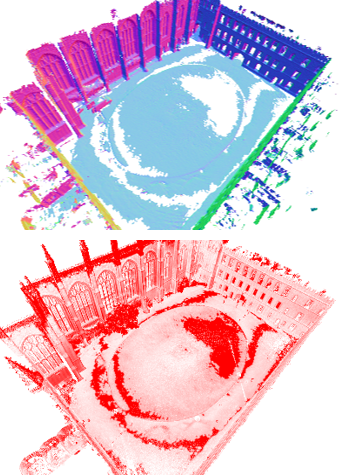}
          }
          \subfigure[Shine With GT pose]{
            \centering
            \includegraphics[width=0.18\textwidth]{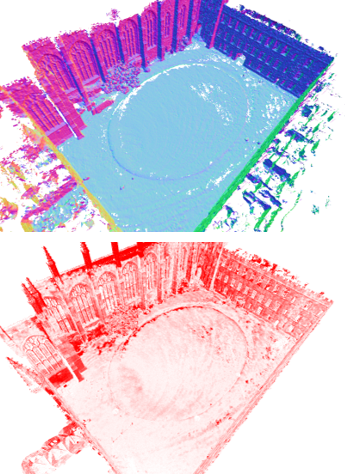}
         }
          \subfigure[Our odomery mapping]{
            \label{fig:odo_map_quli2_d}
             \centering
             \includegraphics[width=0.18\textwidth]{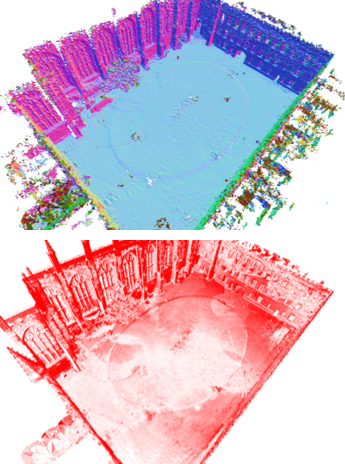}
          }
          \subfigure[Puma odometry mapping]{
            \label{fig:odo_map_quli2_e}
            \centering
            \includegraphics[width=0.18\textwidth]{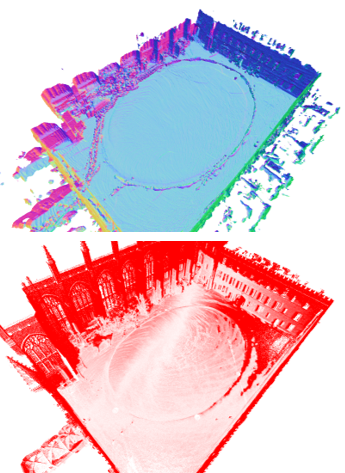}
         }
         \caption{The mapping result with ground truth pose or odometry results on the Newer College~\cite{ramezani2020iros} dataset are shown in the first row. The second  row presents the error maps with ground truth mesh as a reference, where the redder points mean larger error up to 25cm.}
           \vspace{-0.2cm}
         \label{fig:odo_map_quli2}
      \end{figure*}

        Compared to Puma, which involves both odometry and mapping processes, our approach also realizes both odometry and mapping using an implicit neural network and achieves superior performance in almost all metrics. In the MaiCity dataset, the slightly inaccurate trajectory of our method results in a larger distance compared to the completion distance, as also presented in \cref{subsec:trackquality}. However, with a more precise trajectory in the Newer College dataset, our approach significantly outperforms Puma. These results are visually depicted in~\cref{fig:odo_map_quli_d} and \cref{fig:odo_map_quli_e} for MaiCity, and \cref{fig:odo_map_quli2_d} and \cref{fig:odo_map_quli2_e} for Newer College. 
       Although Puma appears more complete, the second row of the figures indicates that this comes at the expense of mapping accuracy. Also, on the Maicity dataset, we can see the ground folds for Puma as it tries to reconstruct a watertight surface and thus is influenced by surrounding objects. As shown on the Newer College dataset, Puma cannot remove the dynamic objects and insufficient points on the wall hinder a complete reconstruction.

      \subsection{Mapping Quality}
      \begin{table*}[t]
         \begin{center}
               \renewcommand\arraystretch{1}
               \setlength{\tabcolsep}{6pt}
               \footnotesize
            \begin{tabular}{lc|cccc|cccc}
               \toprule
               \multirow{2}{*}{Method}& \multirow{2}{*}{Pose} & \multicolumn{4}{c|}{MaiCity}& \multicolumn{4}{c}{Newer College}\\
   
               &&Map. Acc. $\downarrow$&Map. Comp. $\downarrow$&C-l1. $\downarrow$& F-score $\uparrow $&Map. Acc. $\downarrow$&Map. Comp. $\downarrow$&C-l1. $\downarrow$& F-score $\uparrow $\\
               \midrule
               SHINE~\cite{zhong2023icra}&\multirow{3}{*}{GT pose} & 4.17
               & 5.30 &4.74& 89.67 & 8.32
               & \textbf{14.36} &11.34& 90.65\\
               Vdbfusion~\cite{vizzo2022sensors}&&4.12&8.01&6.07& 90.16&6.87&18.37&12.61& 89.96\\
               Ours     &&\textbf{3.15}&\textbf{4.84}&\textbf{4.00}& \textbf{92.96}&\textbf{6.86}&15.59&\textbf{11.24}& \textbf{91.83}\\
               \bottomrule
            \end{tabular}
         \end{center}
           \vspace{-0.1cm}
         \caption{Odometry and Mapping results of the reconstruction quality on
         the MaiCity~\cite{vizzo2021icra} and Newer College~\cite{ramezani2020iros} dataset. The voxel size is 20\,cm and F-score in \% with a 10\,cm error threshold.}
         \label{tab:mapping}
           \vspace{-0.1cm}
      \end{table*}
   
        To eliminate the influence of pose estimation and thoroughly investigate the mapping ability of different methods, we employ ground truth poses to reconstruct the mesh map of the environments. We compare our approach with two pure mapping methods, Shine-Mapping~\cite{zhong2023icra} and VdbFusion~\cite{vizzo2022sensors}, and provide quantitative results in~\cref{tab:mapping}. As can be seen, our approach outperforms all baseline methods across almost all metrics when compared in terms of pure mapping ability. The superiority of our mapping approach is also evident in~\cref{fig:odo_map_quli} and \cref{fig:odo_map_quli2}, where our reconstruction is the most complete, particularly in terms of the ground. The error maps enforce our claims by demonstrating the greater accuracy of our reconstruction. Note that in the Newer College dataset, we reconstruct every five scans, and the results indicate that our mapping process still performs well even with sparse and noisy observations.

   \subsection{Odometry Evaluation}
   \label{subsec:trackquality}
   
      \begin{table}
         \begin{center}
               \renewcommand\arraystretch{1}
               \setlength{\tabcolsep}{10pt}
               \footnotesize
            \begin{tabular}{lcccc}
               \toprule
               Method& Mai00 & Mai01 &NC&KT09\\
               \midrule
               ICP~\cite{besl1992pami}&1.90&0.05& 15.84& 5.86\\
               GICP~\cite{segal2009rss}&1.24&0.13& 1.02&34.25\\
               Puma~\cite{vizzo2021icra}&\textbf{0.25}&0.06& 0.39&\textbf{3.58}\\
               SuMA~\cite{behley2018rss}&2.01&\textbf{0.04}& 1.22&5.00\\
               \midrule
               DeLORA~\cite{nubert2021icra}&57.57&5.12& - &29.09\\
               PWC-LONet~\cite{wang2021cvpr}&3.28&0.09& 15.78&4.60\\
               Ours&1.27&0.13& \textbf{0.15} &4.26\\
               \bottomrule
            \end{tabular}
            
         \end{center}
           \vspace{-0.1cm}
         \caption{RMSE results of odometry. Mai for MaiCity~\cite{vizzo2021icra}, NC for Newer College~\cite{ramezani2020iros}, KT for KITTI~\cite{geiger2012cvpr}, ``-" for failed}
         \label{tab:track_rmse}

      \end{table}
       As discussed, the quality of odometry largely influences the mapping quality. An accurate trajectory can directly improve the reconstruction result and avoid undesired artifacts. Here we present the results of our odometry compared with other non-learning-based and learning-based methods. As mapping methods like Shine-Mapping~\cite{zhong2023icra} and Vdbfusion~\cite{vizzo2022sensors} do not provide pose estimations, they are omitted from the comparison.
       For non-learning-based methods, we compare our odometry results with Puma~\cite{vizzo2021icra}, SuMA~\cite{behley2018rss}, and two registration algorithms based on ICP: point-to-point ICP~\cite{besl1992pami} and generalized-ICP~\cite{segal2009rss}. 
       For learning-based methods, we adopt two SOTA algorithms with code available: DeLORA~\cite{nubert2021icra} and PWC-LONet~\cite{wang2021cvpr}. For other code-unavailable learning-based methods like LO-Net~\cite{li2019cvpr} and DeepPCO~\cite{wang2019iros}, we report their quantitative results from their papers in our supplementary materials (see Sec. C) along with the above-mentioned methods.
       
       We present the RMSE results in~\cref{tab:track_rmse}. Our method achieves comparable results to other methods on the synthetic MaiCity dataset and KITTI09 datasets while achieving the best performance on the Newer College. 
       Notably, our method does not require any pre-training and exhibits strong generalization ability across different datasets, while pre-trained methods such as DeLORA and PWC-LONet, which are pre-trained on the KITTI dataset, exhibit worse performance on other datasets. Although PWC-LONet still obtains acceptable results on the MaiCity dataset, it almost fails on the Newer College dataset. More results on KITTI can be found in the supplementary materials (see Sec. C).

   \subsection{Ablation Study}
   \label{subsect:abla}
   
      \begin{table}[t]
         \begin{center}
               \renewcommand\arraystretch{1}
               \setlength{\tabcolsep}{2.8pt}
               \footnotesize

            \begin{tabular}{c|cc|ccccc}
               \toprule
               Dataset & Ground & KF-ref.& RMSE$\downarrow$ & Acc.$\downarrow$& Comp.$\downarrow$& C-l1.$\downarrow$ & F$\uparrow $\\
               \midrule
               \multirow{4}{*}{MaiCity}&\ding{55}  & \ding{55}& 0.20 & 6.15& 69.64 & 37.90 & 49.39\\
               &\ding{55}  & \checkmark & 0.20 & 6.13& 70.48 & 38.30 & 48.78\\
               &\checkmark & \ding{55} & 0.17 & 5.93& 11.49 & 8.71 & 76.15\\
               &\checkmark  & \checkmark  & 0.17 & 5.69& 11.23 & 8.46 & 77.26\\
               \hline
               \multirow{4}{1cm}{Newer College}&\ding{55} & \ding{55} & - & -& - & - & -\\
               &\ding{55}  & \checkmark & - & -& - & - & -\\
               &\checkmark & \ding{55}& 0.15 & 16.41& 25.75 & 21.08 & 61.10\\
               &\checkmark  & \checkmark  & 0.15 & 12.89& 22.21 & 17.55 & 74.37\\
               \bottomrule
            \end{tabular}
         \end{center}
           \vspace{-0.1cm}
         \caption{Ablation study of our designs on Maicity~\cite{vizzo2021icra}, Newer College~\cite{ramezani2020iros}. ``-" stands for failed}
         \label{tab:ablation}

      \end{table}

      \begin{figure}[t]
         \subfigure[Map with ground separation]{
            \centering
            \includegraphics[width=0.22\textwidth]{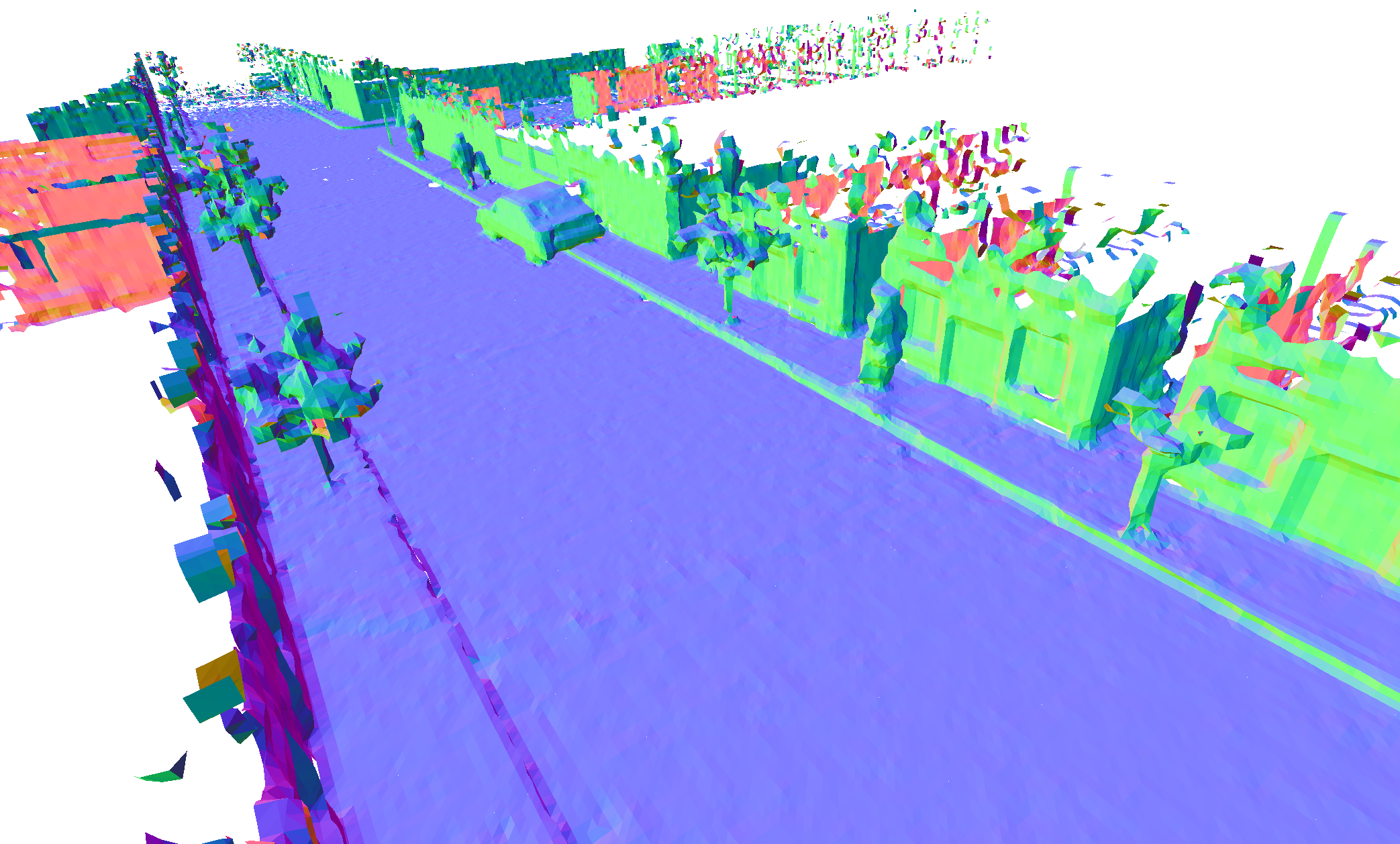}
         }
         \subfigure[Map with ground separation]{
            \centering
             \includegraphics[width=0.22\textwidth]{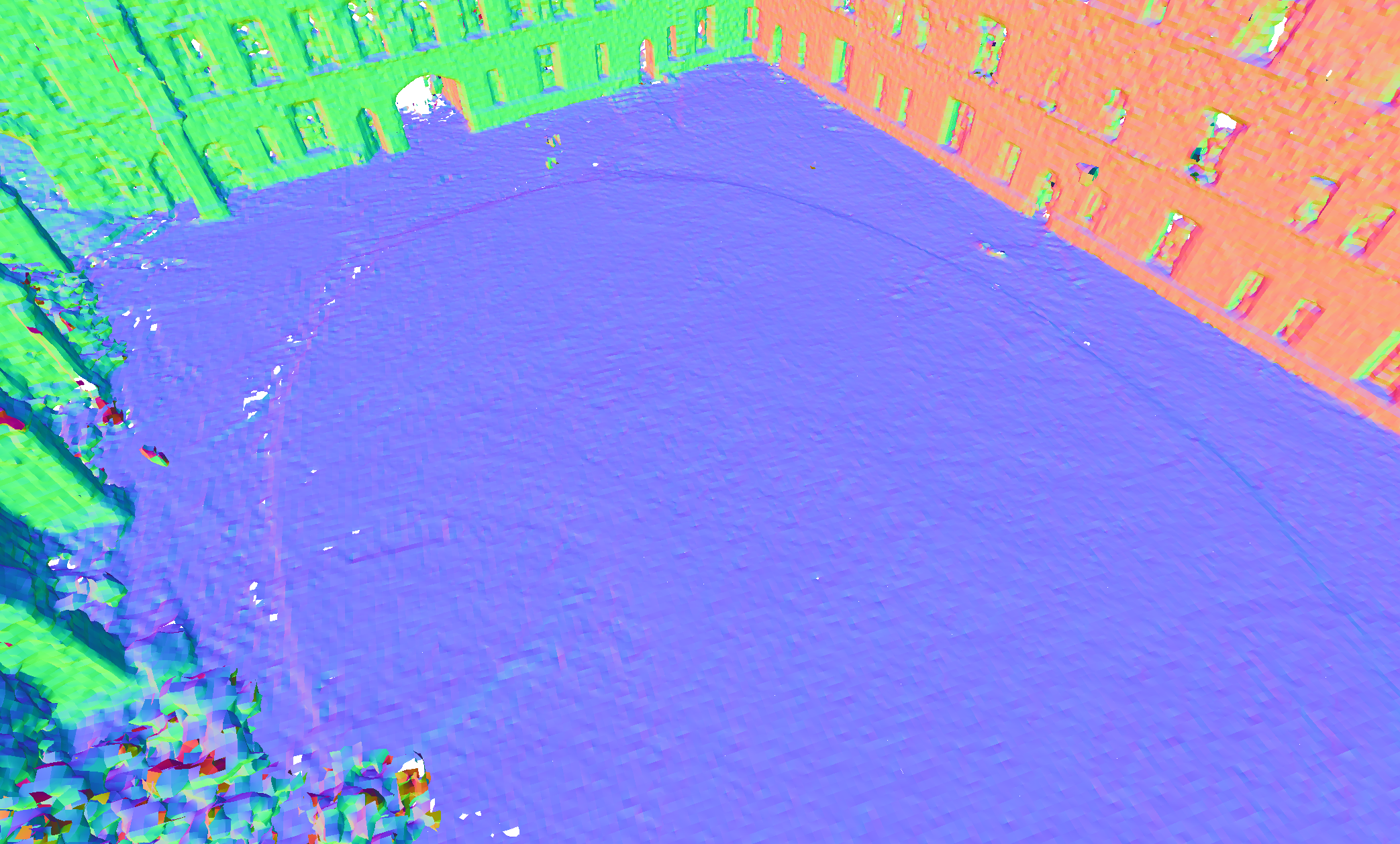}
          }
   
          \subfigure[Map w/o ground separation]{
            \centering
            \includegraphics[width=0.22\textwidth]{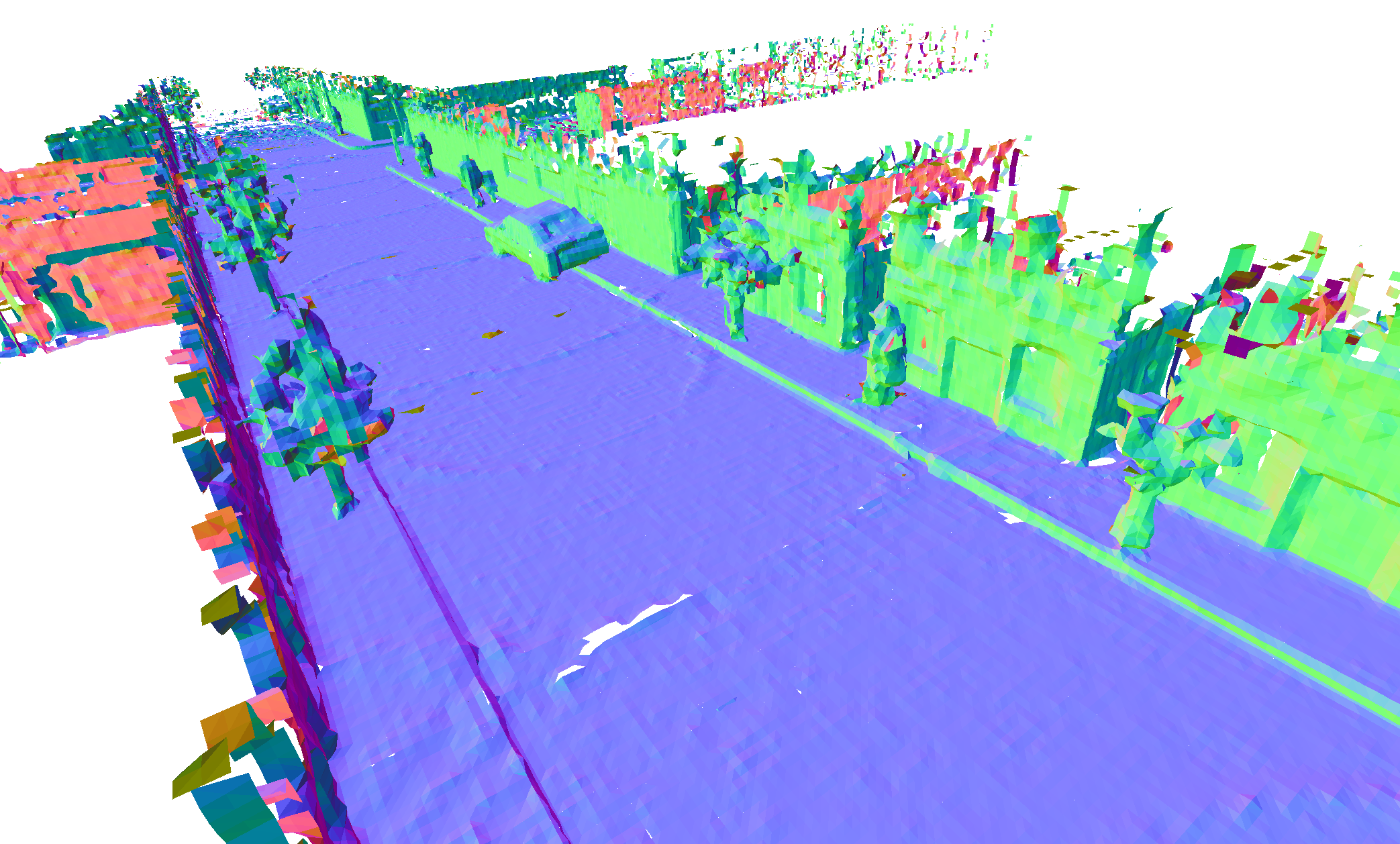}
         }
          \subfigure[Map w/o ground separation]{
             \centering
             \includegraphics[width=0.22\textwidth]{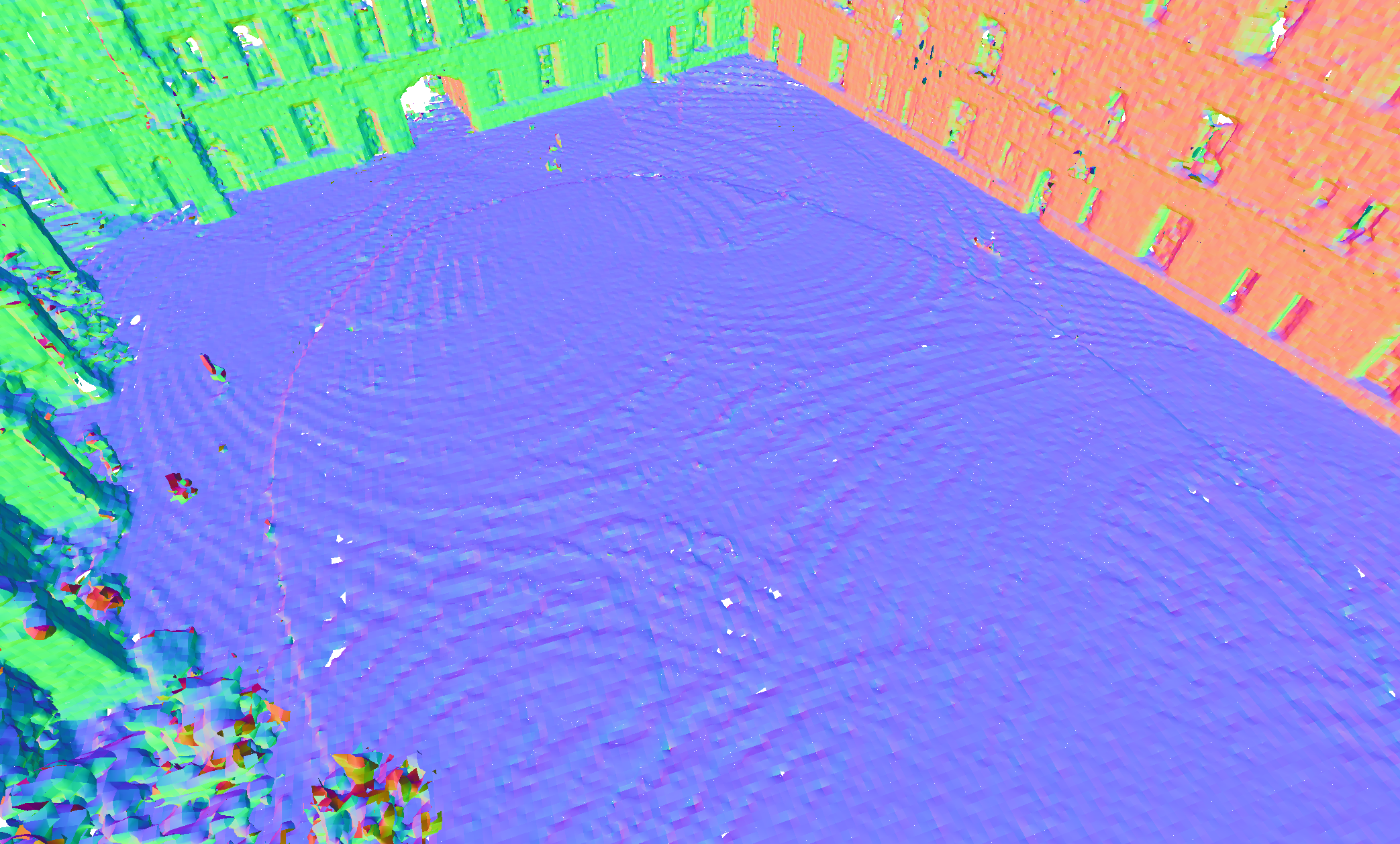}
          }
         \caption{Ablation study for ground separation in mapping using the pose provided by our neural odometry. With ground separation, the mapping result is neater and completer.}
         \label{fig:abla_patch_map}
           \vspace{-0.4cm}
      \end{figure}
   
      {\bf Ground separation.} 
           We compare the performance of our method with and without ground separation and show the odometry and mapping accuracy in~\cref{tab:ablation}.  For odometry accuracy, we see that RMSE error declines with ground separation for the MaiCity dataset and for Newer College, the approach even failed without ground separation. Moreover, when checking pose error in each axis (supplementary materials Sec. D), the trajectory with ground separation is consistent in the z-axis, while without ground separation, it diverges fast. For mapping accuracy, all mapping metrics indicate that our method achieves significantly better mapping results with ground separation. We can also see a clear improvement visually in~\cref{fig:abla_patch_map}. With ground separation, the ``ripples effect" is suppressed and the holes are disappeared.
       
      \begin{figure}[t]
            \centering
         \subfigure[Mapping with key-scan refine]{
            \centering
            \includegraphics[width=0.22\textwidth]{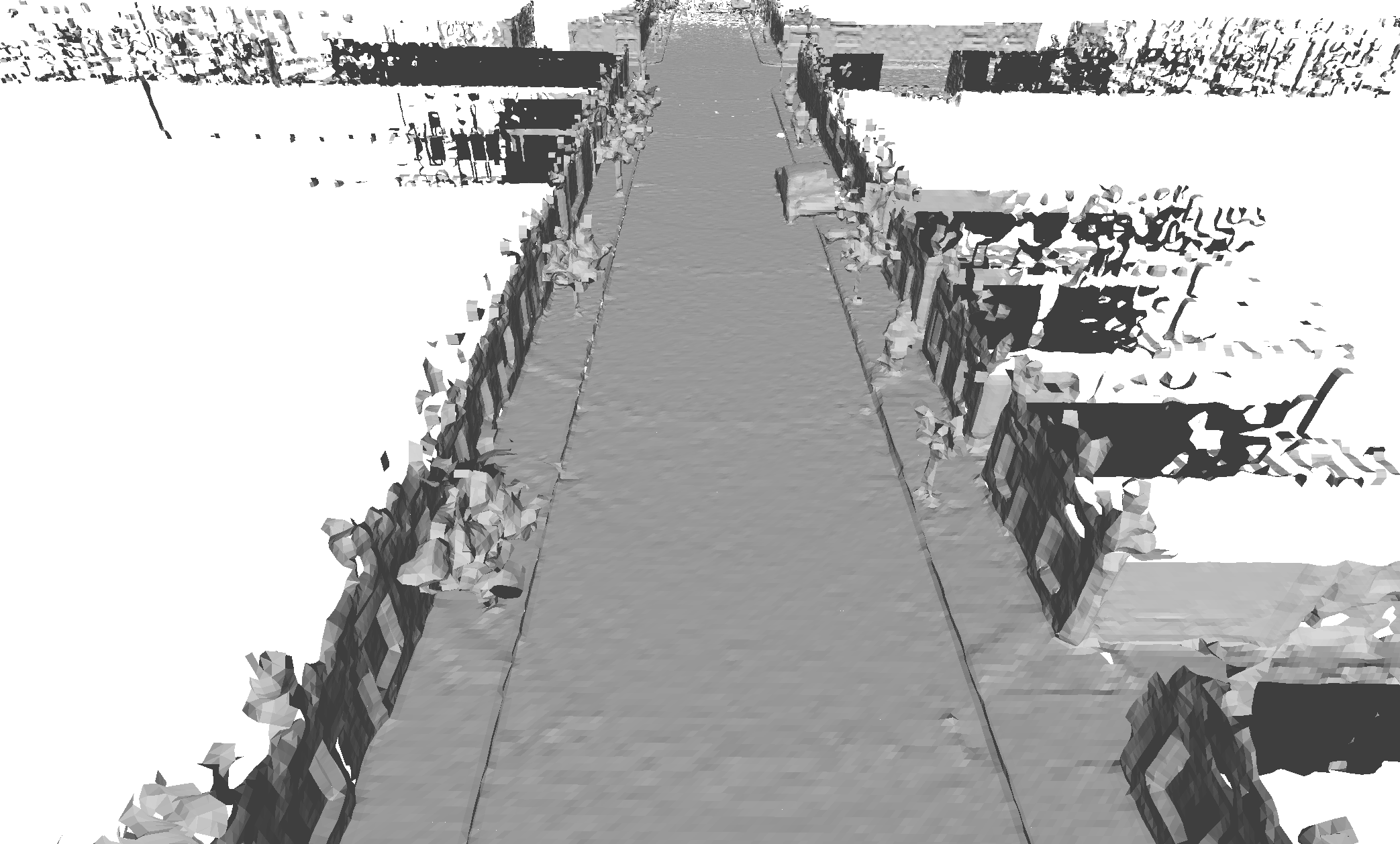}
         }
         \subfigure[Mapping with key-scan refine]{
            \centering
             \includegraphics[width=0.22\textwidth]{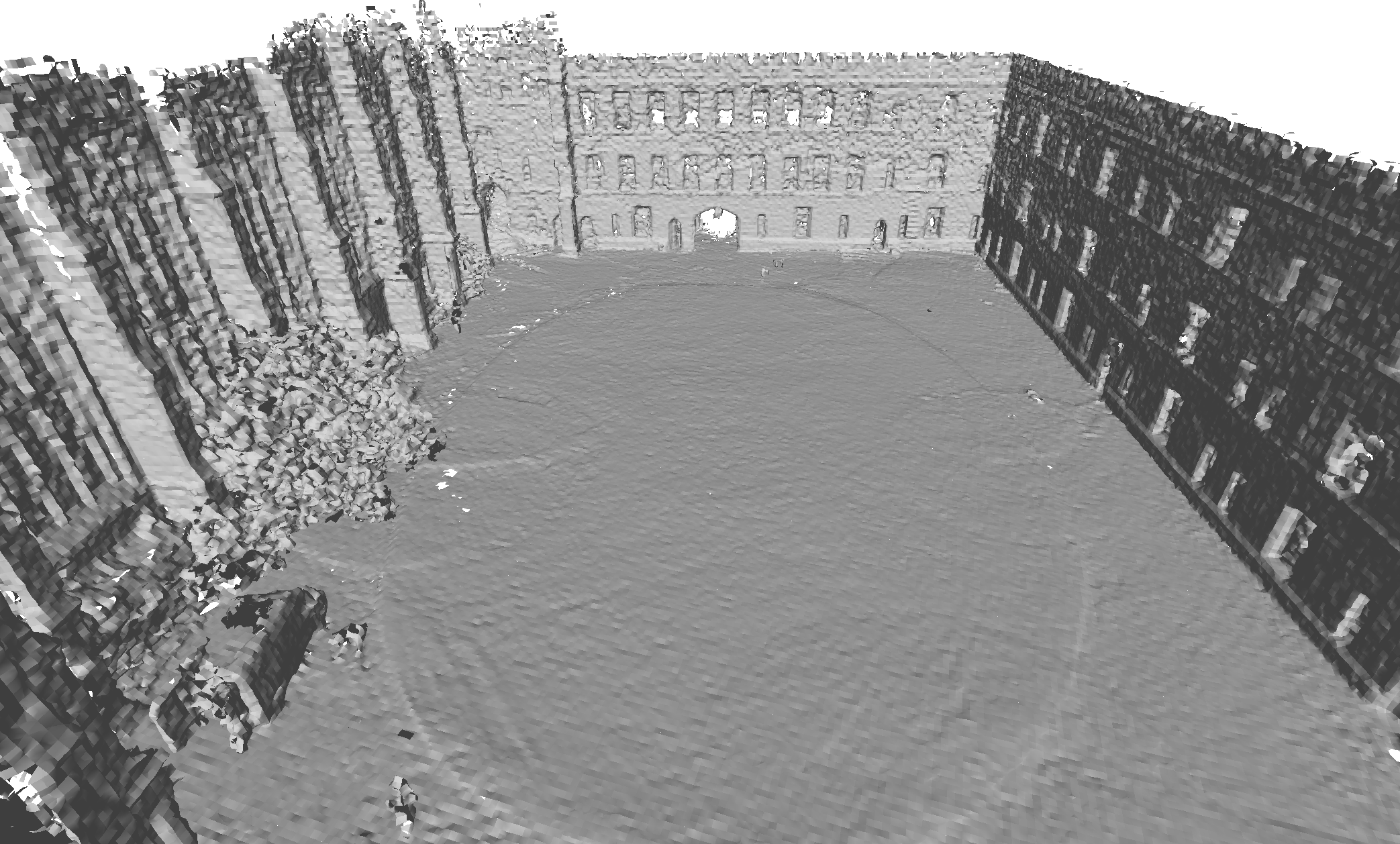}
          }
         \subfigure[Mapping w/o key-scan refine]{
            \centering
            \includegraphics[width=0.22\textwidth]{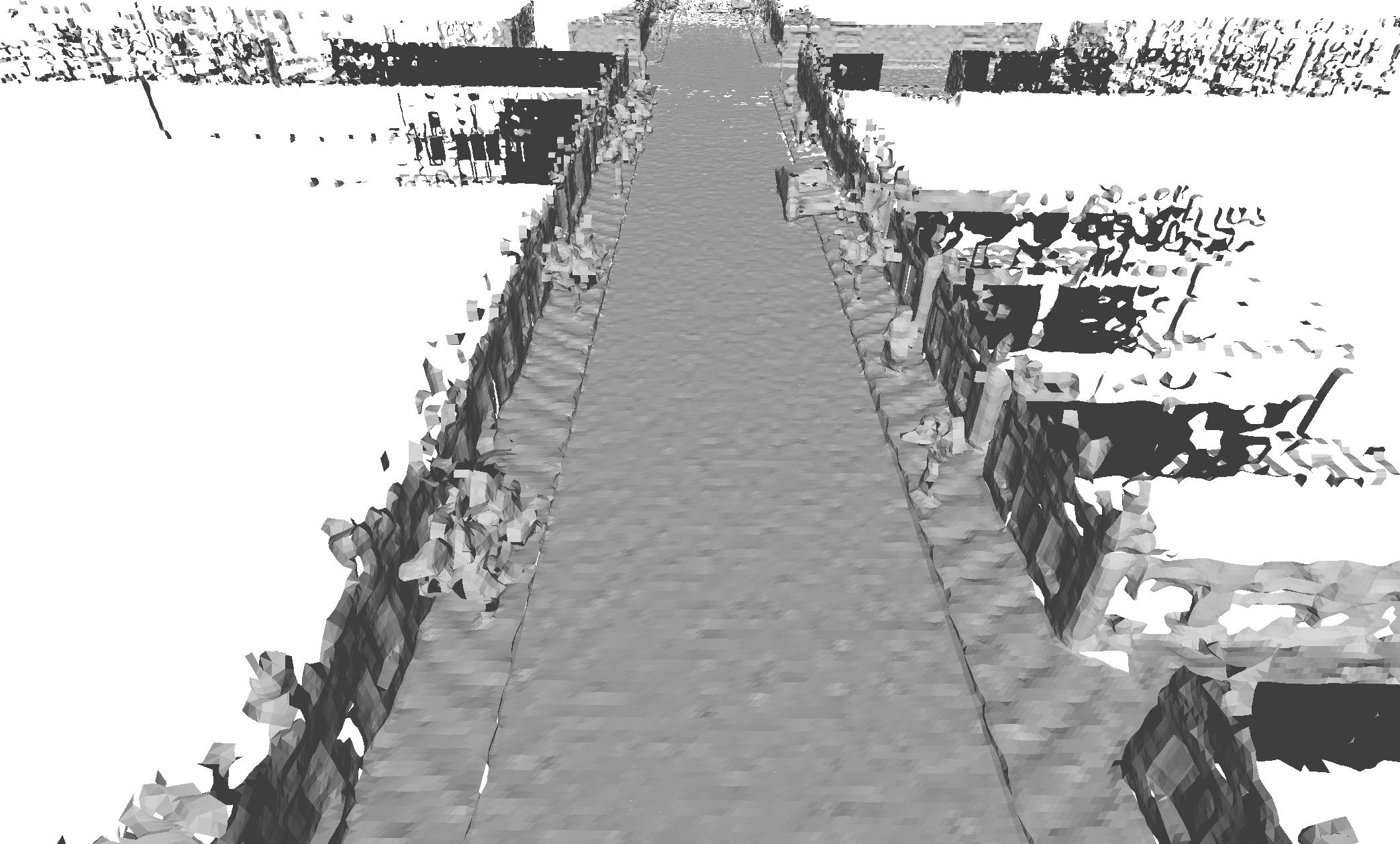}
         }
          \subfigure[Mapping w/o key-scan refine]{
             \centering
             \includegraphics[width=0.22\textwidth]{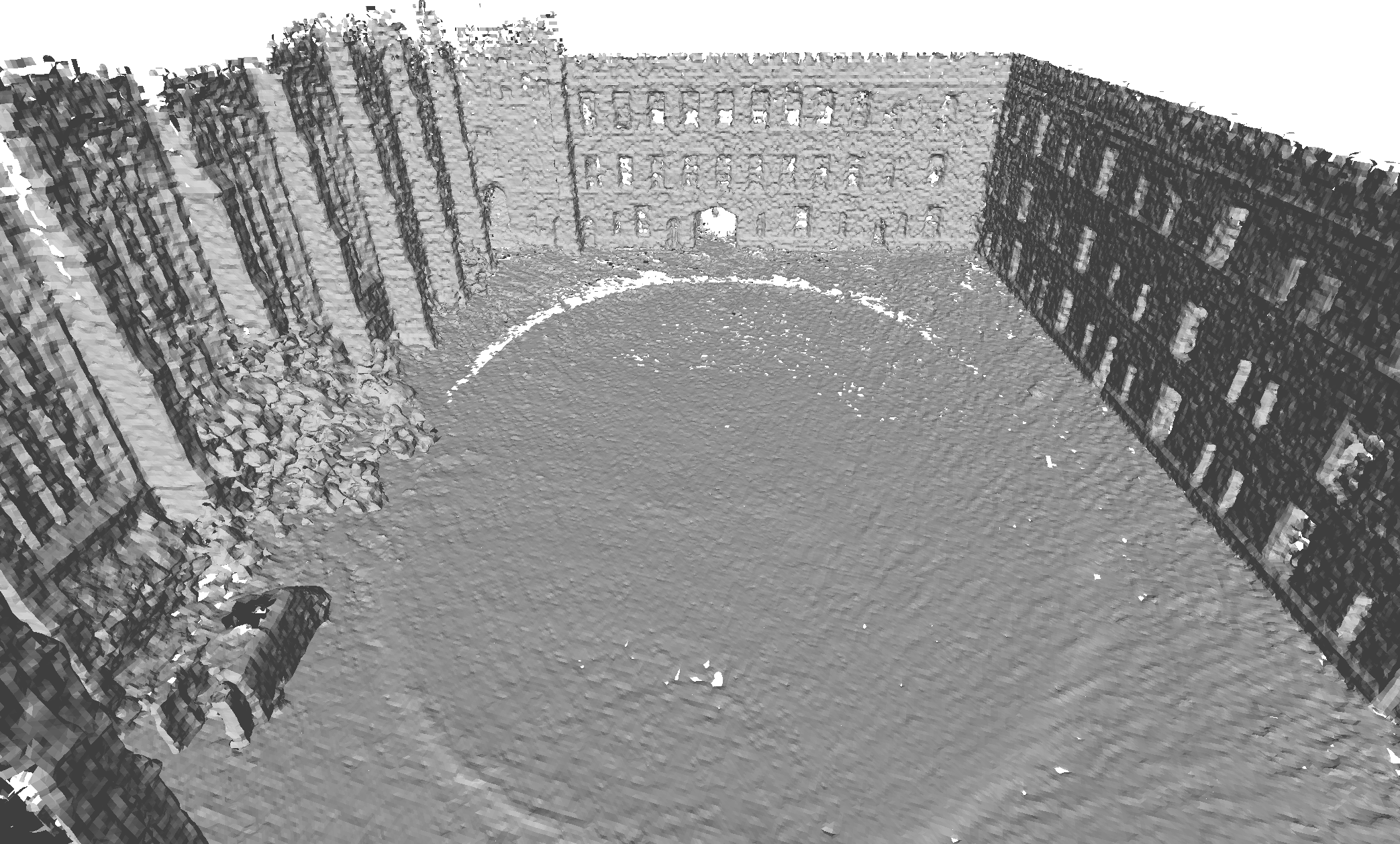}
          }
         \caption{Ablation study for key-scan refine in terms of mapping. The pose is obtained by the SLAM odometry. The key-scan refine makes the reconstruction result clearer.}
         \label{fig:abla_key_map}
           \vspace{-0.2cm}
      \end{figure}
   
       {\bf Key-scan refine strategy.} 
       We further analyze the effectiveness of our key-scan refine strategy and show the result in \cref{tab:ablation}. The numerical results show improvement with key-scan refinement, and the visual improvement is even more significant, as shown~\cref{fig:abla_key_map}. The key-scan refinement produces smoother and more complete results, as evidenced by the improved maps of roads, walls, and vehicles.
      
      \begin{figure}[t]
         \begin{center}
               \vspace{-0.5cm}

             \subfigure[Time vs Acc. on Newer College]{
                \centering
                \includegraphics[width=0.225\textwidth]{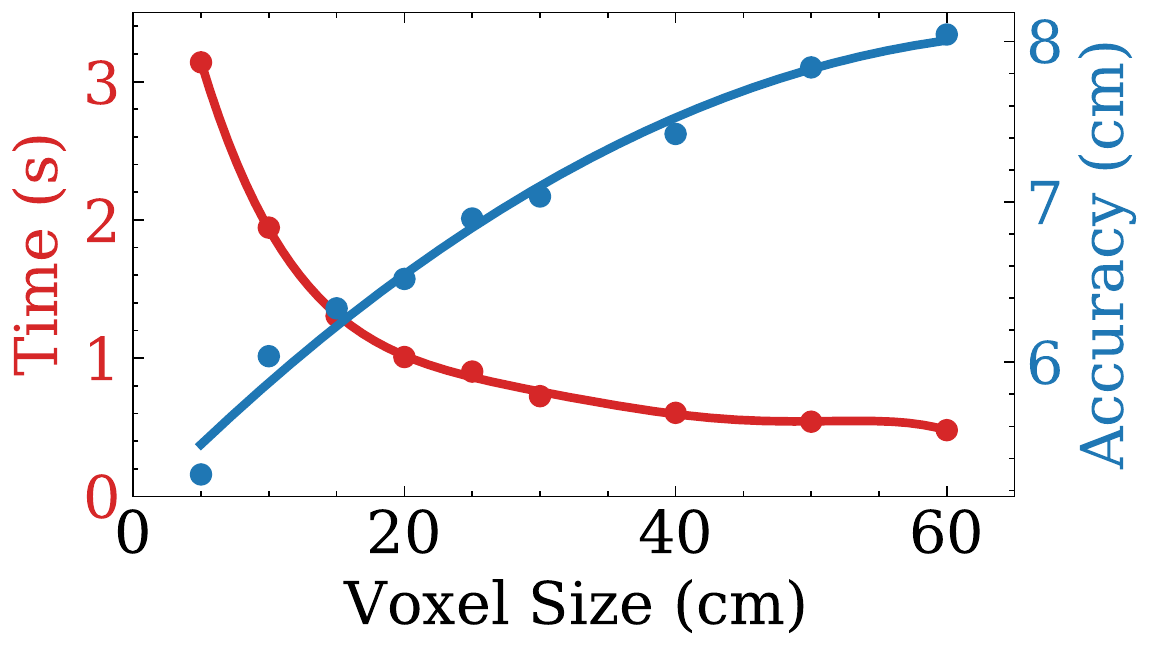}
             }
            \subfigure[Me. vs Acc. on Newer College]{
            \centering
            \includegraphics[width=0.225\textwidth]{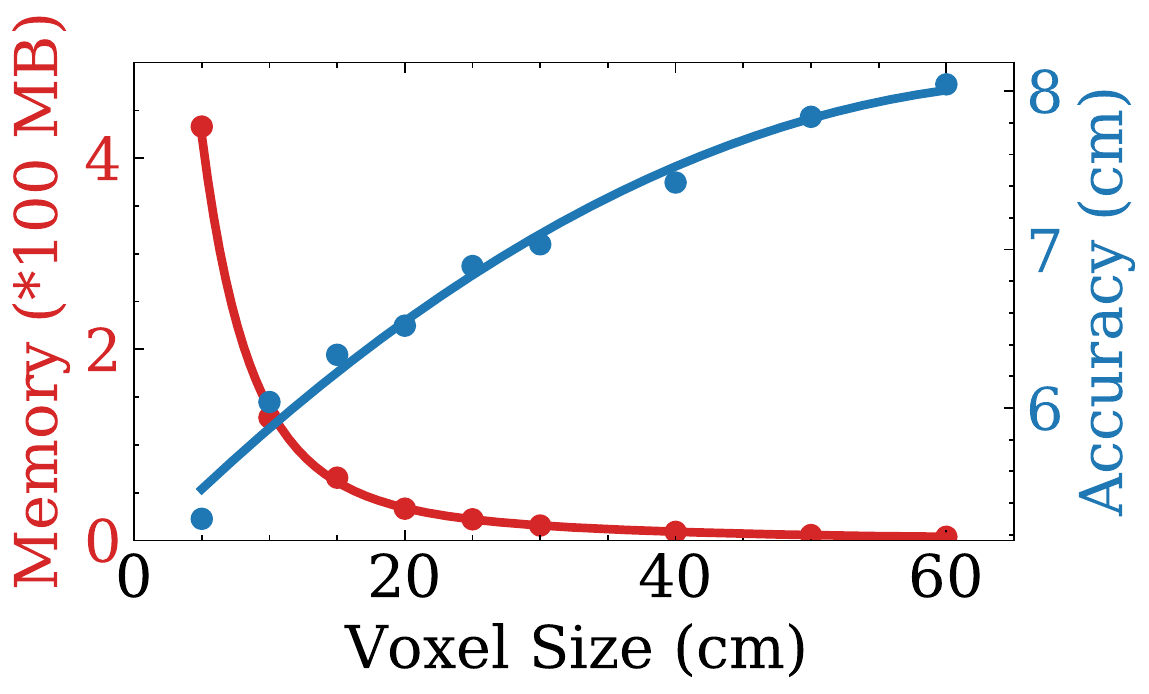}
            }
         \end{center}
         \caption{Study on voxel size v.s. processing time, memory consumption and accuracy distance on Newer College~\cite{ramezani2020iros}.}
         \label{fig:ablavoxsize}
           \vspace{-0.4cm}
      \end{figure}
   
       {\bf Voxel size.} 
       We analyze the mapping quality, memory consumption, and processing time v.s. the voxel size shown in~\cref{fig:ablavoxsize}. We text our NeRF-LOAM on an Intel Xeon CPU with 2.1 GHz and an Nvidia NVIDIA Titan RTX with 24 GB of memory. The results show that the mapping performance decreases as the voxel size exceeds 20\,cm, while processing time and memory consumption remain constant. Thus, we set the voxel size as 20\,cm.
       More studies on parameters are provided in the appendix (see Sec. D).

   \section{Conclusion}
      In this paper, we presented a novel approach for simultaneous odometry and mapping using neural implicit representation with 3D LiDAR data. The devised NeRF-LOAM network tackles incremental LiDAR inputs in outdoor large-scale environments. It uses voxel embeddings to record the geometrical structure and avoids any pre-training, thus generalizing well in different situations. We further conceive a dynamic embedding generation, which realizes fast query and allocation to support outdoor large-scale applications. Experiments conducted on simulated and real-world datasets showed that our approach reconstructs higher-quality 3D mesh maps compared to other learning-based or non-learning-based methods. Our method estimates at the same time an accurate pose and generalizes well without any offline pre-training.
    
       \textbf{Limitation and future work.} 
       Our NeRF-LOAM cannot currently operate in real-time with our unoptimized Python implementation. The primary bottleneck is the intersection query between the ray and the map. For future work, we can facilitate the runtime by using sliding windows or local searching based on the estimated odometry pose and optimize the code in C++. Additionally, we plan to combine our work with loop closures to handle drift in long-term tracking and mapping, ultimately achieving a full SLAM system.
   
       \textbf{Societal Impacts.} Our approach provides accurate trajectories and reconstructs a dense environmental awareness map. This is particularly important for safety-critical real-world applications, such as autonomous cars.

      \newpage
      {\small
         \bibliographystyle{ieee_fullname}
         \bibliography{deng23}
      }
      
      \twocolumn[
         \begin{@twocolumnfalse}
            \section*{\centering{NeRF-LOAM: Neural Implicit Representation for Large-Scale  \\ Incremental LiDAR Odometry and Mapping Supplementary Material}}
            \centering
         \end{@twocolumnfalse}
         ]
         \setcounter{section}{0}
         \renewcommand\thesection{\Alph{section}}
         \section{ Simultaneously Odometry \& Mapping Results}
         \label{supp:sec:odomap}

\begin{figure}[htbp]
            \subfigure[Ground Truth point cloud map]{
               \centering
               \includegraphics[width=0.45\textwidth]{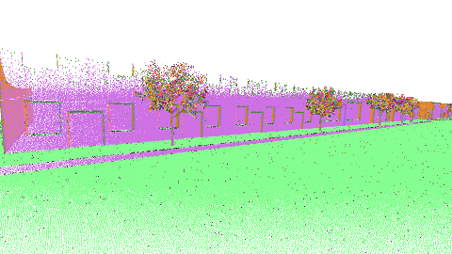}
            }
      
            \subfigure[Ours Odometry mapping result with error map]{
               \centering
               \includegraphics[width=0.45\textwidth]{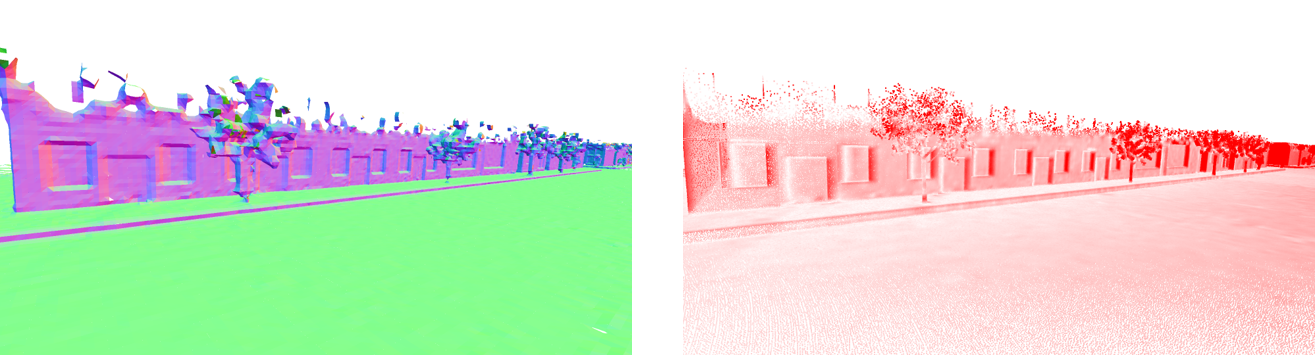}
            }
            \subfigure[Puma Odometry mapping result with error map]{
               \centering
               \includegraphics[width=0.45\textwidth]{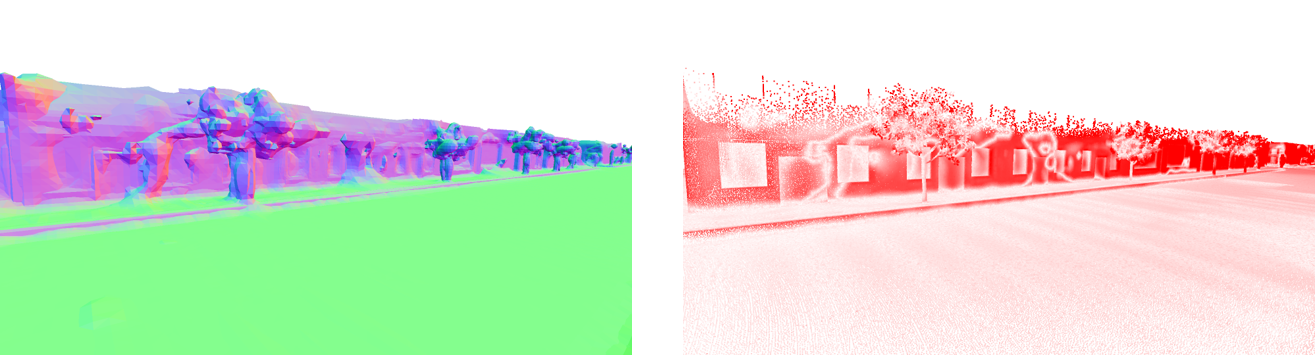}
            }
      
            \caption{The simultaneously odometry \& mapping results with error maps on the MaiCity~\cite{vizzo2021icra} dataset. The error maps are with the ground truth map as a reference, where the redder points mean larger errors up to 20\,cm.}
            \label{fig:supp_odo_map_quli}
            \vspace{-0.1cm}
         \end{figure}

\begin{figure}[htbp]
            \subfigure[Ground Truth point cloud map]{
               \centering
               \includegraphics[width=0.45\textwidth]{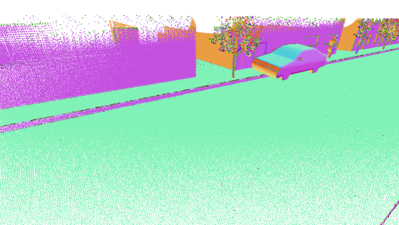}
            }
      
            \subfigure[Ours Odometry mapping result with error map]{
               \centering
               \includegraphics[width=0.45\textwidth]{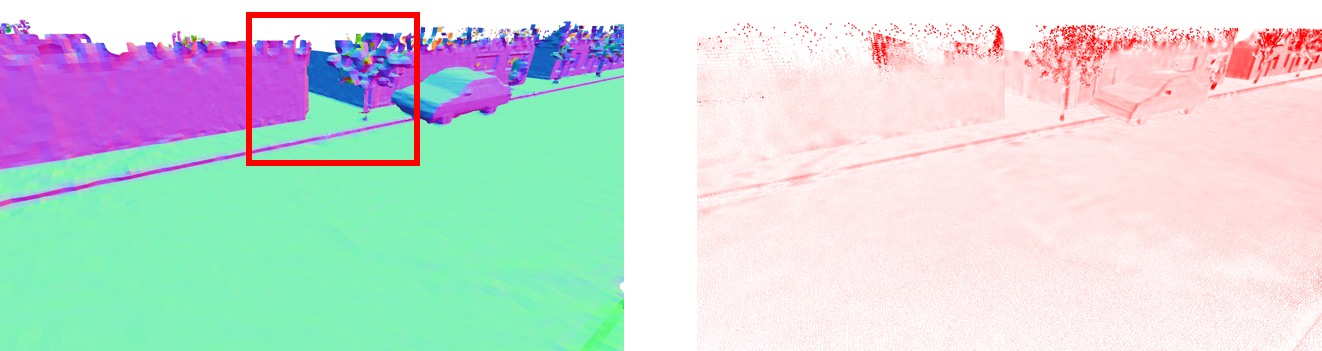}
            }
            \subfigure[Puma Odometry mapping result with error map]{
               \centering
               \includegraphics[width=0.45\textwidth]{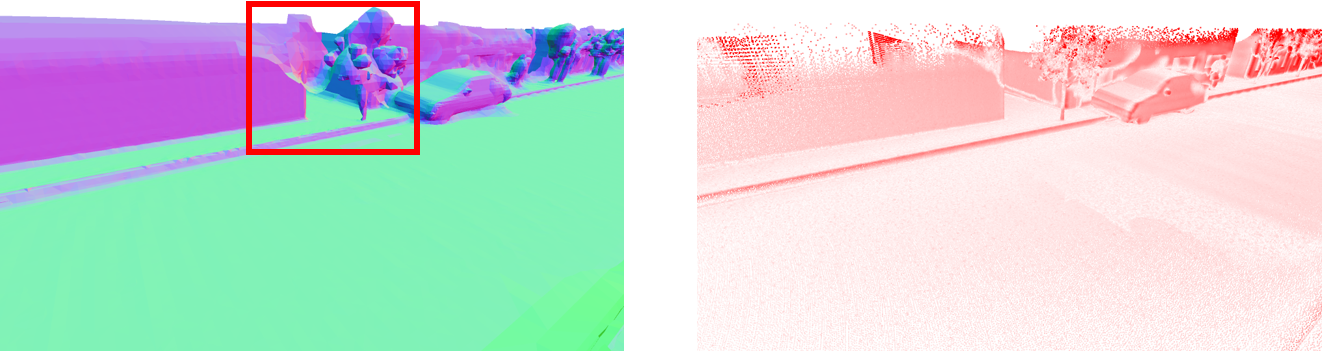}
            }
      
            \caption{The simultaneously odometry \& mapping results with error maps on the MaiCity~\cite{vizzo2021icra} dataset. The error maps are with the ground truth map as a reference, where the redder points mean larger errors up to 20\,cm. The red box illustrates the unreal reconstruction.}
            \label{fig:supp_odo_map_quli2}
            \vspace{-0.1cm}
         \end{figure}

         We present in ~\cref{fig:supp_odo_map_quli} and ~\cref{fig:supp_odo_map_quli2} our odometry mapping results compared with Puma~\cite{vizzo2021icra} herewith the corresponding ground truth map.  To reconstruct more complete results, Puma uses polynomial function and normals to simulate the surrounding meshes, resulting in loss of detail (e.g., windows, curb) and unreal reconstruction (e.g., the connection of tree and wall), while our reconstruction can provide higher accuracy and neater results.
         
\begin{figure}[htbp]
            \subfigure[Ground Truth point cloud map]{
               \centering
               \includegraphics[width=0.225\textwidth]{figures/supp_odomap_00.png}
            }
            \subfigure[Ours with KissICP]{
               \centering
               \includegraphics[width=0.225\textwidth]{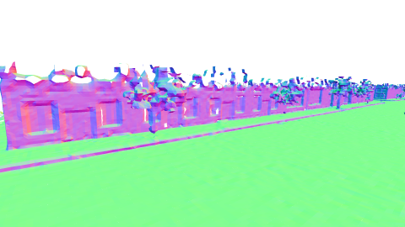}
             }
      
             \subfigure[Vdbfusion with KissICP]{
               \centering
               \includegraphics[width=0.225\textwidth]{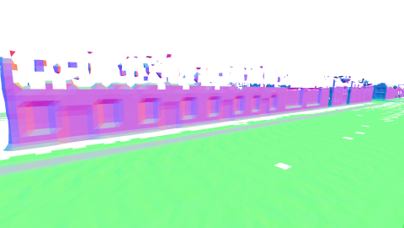}
            }
            \subfigure[SHINE-Mapping with KissICP]{
               
                \centering
                \includegraphics[width=0.225\textwidth]{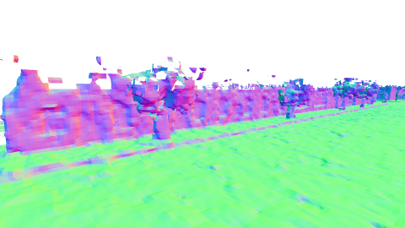}
             }
      
            \caption{The mapping results combined with KissICP\cite{vizzo2023ral} odometry on the MaiCity~\cite{vizzo2021icra} dataset.}
            \label{fig:supp_odo_map_quli3}
            \vspace{-0.1cm}
         \end{figure}

\begin{figure}[htbp]
            \subfigure[Ground Truth point cloud map]{
               \centering
               \includegraphics[width=0.225\textwidth]{figures/supp_odomap_06.png}
            }
            \subfigure[Ours  with KissICP]{
               \centering
               \includegraphics[width=0.225\textwidth]{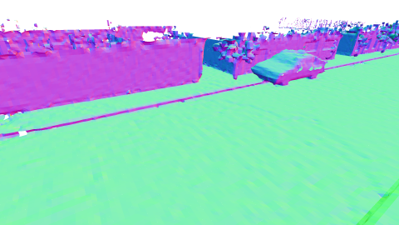}
             }
      
             \subfigure[Vdbfusion with KissICP]{
               \centering
               \includegraphics[width=0.225\textwidth]{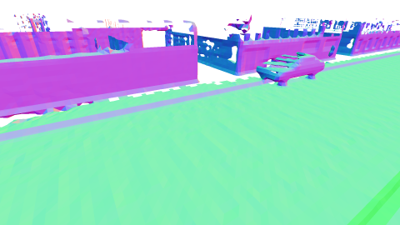}
            }
            \subfigure[SHINE-Mapping with KissICP]{
               
                \centering
                \includegraphics[width=0.225\textwidth]{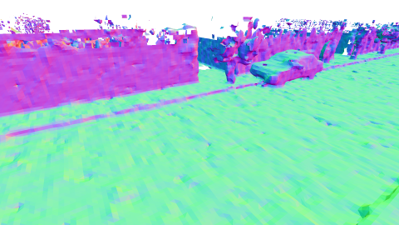}
             }
      
             \caption{The mapping result combined with KissICP\cite{vizzo2023ral} odometry on the MaiCity~\cite{vizzo2021icra} dataset.}
            \label{fig:supp_odo_map_quli4}

         \end{figure}

         We also present in ~\cref{fig:supp_odo_map_quli3} and ~\cref{fig:supp_odo_map_quli4} the mapping results on MaiCity dataset of SHINE-Mapping~\cite{zhong2023icra} and Vdbfusion~\cite{vizzo2022sensors} combined with KissICP~\cite{vizzo2023ral} odometry. These results also demonstrate that ours can provide a complete and smooth map.

         \begin{table*}[htbp]
         \begin{center}
            \resizebox{\linewidth}{!}{
               \begin{tabular}{c||cc|cc|cc||cc|cc|cc|cc|cc}
                  \toprule
                  \multirow{2}{*}{Method} & \multicolumn{2}{c|}{00} & \multicolumn{2}{c|}{01}& \multicolumn{2}{c|}{03}&\multicolumn{2}{c|}{04}&\multicolumn{2}{c|}{05}&\multicolumn{2}{c|}{07}&\multicolumn{2}{c|}{09}&\multicolumn{2}{c}{10}\\
                  \cmidrule{2-17}
                  & $t_{rel}$&$r_{rel}$& $t_{rel}$&$r_{rel}$& $t_{rel}$&$r_{rel}$&$t_{rel}$&$r_{rel}$&$t_{rel}$&$r_{rel}$&$t_{rel}$&$r_{rel}$&$t_{rel}$&$r_{rel}$&$t_{rel}$&$r_{rel}$\\
                  \bottomrule
                  \toprule
                  ICP-po2po~\cite{besl1992pami}&6.88&2.99&11.21&2.58&11.07&5.05& 6.64&4.05&3.97&1.93&5.17&3.35&6.93&2.89&8.91&4.47\\
                  ICP-po2pl~\cite{Rusinkiewicz20013di}&3.80&1.73&13.53&2.58&2.72&1.63&2.96&2.58&1.55&1.42&1.55&1.42&3.95&1.71&6.13&2.60\\
                  GICP~\cite{segal2009rss}&\textbf{1.29}&0.64&4.39&2.58&1.68&1.08& 3.76&1.93&1.02&0.54&0.64&0.46&1.97&0.77&\textbf{1.31}&\textbf{0.62}\\
                  SUMA~\cite{behley2018rss}&2.93&0.92&4.05&1.22&1.43&\textbf{0.75}& 11.90&1.06 &1.46&0.79&1.75&1.17&1.92&0.78&1.81&0.97\\
                  PUMA(NN)\cite{vizzo2021icra}&2.15&1.14&4.32&1.04&\textbf{1.34}&1.07& 2.09&1.46 &1.56&1.07&1.88&1.36&1.80&0.82&2.24&1.67\\
                  PUMA(RC)\cite{vizzo2021icra}&1.55&0.74&3.38&1.00&1.60&1.10& \textbf{1.63}&\textbf{0.92} &\textbf{1.20}&\textbf{0.61}&0.72&0.55&1.51&0.66&1.38&0.84\\
                  
                  \bottomrule
                  \toprule
                  DeLORA~\cite{nubert2021icra}&-&-&-&-&-&-&-&-&-&-&-&-&9.07&3.14&6.53&4.22\\
                  DeepPCO~\cite{wang2019iros}&-&-&-&-&-&-&2.63&3.05&-&-&-&-&-&-&2.21&1.67\\
                  LONet~\cite{li2019cvpr}&1.47*&0.72*&1.36*&0.47*&1.03*&0.66*&0.51*&0.64*&1.04*&0.69*&1.70&0.89&1.37&0.58&1.80&0.93\\
                  PWCLONet~\cite{wang2021cvpr}&0.78*&0.42*&0.67*&0.23*&0.76*&0.44*&0.37*&0.40*&0.45*&0.27*&\textbf{0.60}&\textbf{0.44}&\textbf{0.79}&\textbf{0.35}&1.69&\textbf{0.62}\\
                  Ours&1.34&\textbf{0.54}&\textbf{2.07}&\textbf{0.52}&2.22&1.57& 1.74&1.00&1.40&0.65&1.00&0.65&1.63&0.57&2.08&0.69 \\
                  \bottomrule
               \end{tabular}
            }
         \end{center}
         \caption{The odometry results on KITTI dataset~\cite{geiger2012cvpr}. $t_{rel}$ and $r_{rel}$ mean the average translational RMSE (\%) and rotational RMSE ($^\circ$/100\,m) respectively on all possible subsequences in the length of 100, 200, ..., 800\,m. ``*" means the results on the training sequence, ``-" means not provided, RC for ray casting, NN for nearest neighbor. The best results are bold whereas the results on training sequences are beyond our consideration.} 
         \label{tab:track_kitti}
         \vspace{-0.4cm}
      \end{table*}
      
\begin{figure*}[htbp]
            \begin{center}
               \includegraphics[width=\textwidth]{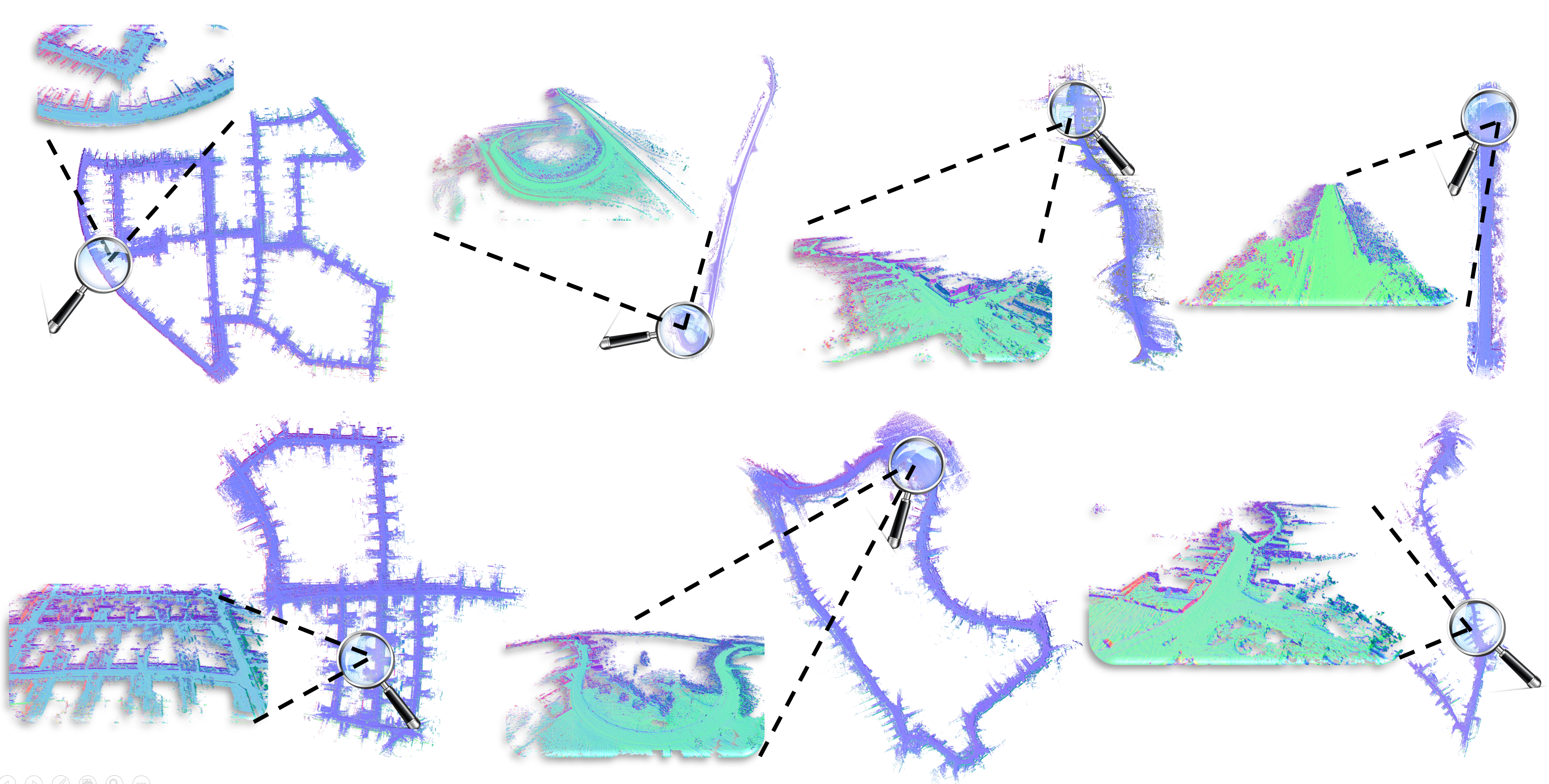}
            \end{center}
            \caption{The qualitative result of our odometry mapping on KITTI~\cite{geiger2012cvpr} dataset. From left upper to right bottom, we list the results of sequences 00, 01, 03, 04, 05, 09, 10}
            \label{fig:supp_odo_map_kitti5}

         \end{figure*}

         To demonstrate our odometry and mapping under large-scale environments. We provide in ~\cref{fig:supp_odo_map_kitti5} the qualitative results of odometry mapping on the KITTI~\cite{geiger2012cvpr} dataset. Our method can obtain a fine reconstruction of city environments without a loop. However, for trajectories with a loop, our method cannot maintain a consistent global map.

         \section{Mapping Quality}
         
         The ground truth pose is used in this section to compare our mapping ability with SHINE-Mapping~\cite{zhong2023icra} and Vdbfusion~\cite{vizzo2022sensors}. As the detailed reconstruction results are similar to the results of ~\cref{supp:sec:odomap}, we provide in ~\cref{{fig:supp_map_birdview}} the bird-eye view of reconstruction on Maicity~\cite{vizzo2021icra} dataset. As we can see, SHINE-Mapping provides a relatively complete map but is not smooth enough. While Vdbfusion provides the smoothest map but the map is not complete. Our mapping process can provide the most complete and smooth result.

         \begin{table}[H]
            \begin{center}
               \renewcommand\arraystretch{1}
               \setlength{\tabcolsep}{2.8pt}
               \footnotesize

               \begin{tabular}{c|ccc|ccccc}
                  \toprule
                  Dataset & Grd & KS&GT&RMSE $\downarrow$ &Acc. $\downarrow$&Comp. $\downarrow$&C-l1. $\downarrow$& F$\uparrow $\\
                  \midrule
                  \multirow{8}{*}{MaiCity}&\ding{55}  & \ding{55}&\ding{55} & 0.20 & 6.15& 69.64 & 37.90 & 49.39\\
                  &\ding{55}  & \checkmark&\ding{55} & 0.20 & 6.13& 70.48 & 38.30 & 48.78\\
                  &\checkmark & \ding{55}&\ding{55} & 0.17 & 5.93& 11.49 & 8.71 & 76.15\\
                  &\checkmark  & \checkmark&\ding{55}  & 0.17 & 5.69& 11.23 & 8.46 & 77.26\\
                  &\ding{55}  & \ding{55}&\checkmark & - & 3.57& 5.61 & 4.59 & 90.61\\
                  &\ding{55}  & \checkmark&\checkmark & - & 3.43& 5.40 & 4.42 & 90.81\\
                  &\checkmark & \ding{55}&\checkmark    & - & 3.27& 5.03 & 4.15 & 92.80\\
                  &\checkmark  & \checkmark&\checkmark  & - & 3.15& 4.84& 4.00 & 92.96\\
                  \hline
                  \multirow{8}{1cm}{Newer College}&\ding{55} & \ding{55}&\ding{55} & \multirow{2}{*}{Failed} & -& - & - & -\\
                  &\ding{55}  & \checkmark&\ding{55} & & -& - & - & -\\
                  &\checkmark & \ding{55}&\ding{55} & 0.15 & 16.41& 25.75 & 21.08 & 61.10\\
                  &\checkmark  & \checkmark&\ding{55}  & 0.15 & 12.89& 22.21 & 17.55 & 74.37\\
                  &\ding{55}  & \ding{55}&\checkmark & - & 7.01& 15.58 & 11.29 & 91.58\\
                  &\ding{55}  & \checkmark&\checkmark & - &6.73& 14.86 & 10.79 & 91.92\\
                  &\checkmark & \ding{55}&\checkmark & - & 7.50& 16.75 & 12.13 & 90.98\\
                  &\checkmark  & \checkmark&\checkmark  & - & 6.86& 15.62 & 11.24 & 91.84\\
                  \bottomrule
               \end{tabular}
            \end{center}
            \caption{Ablation study of our designs on Maicity~\cite{vizzo2021icra}, Newer College~\cite{ramezani2020iros}. ``-" stands for no meaning data. "Grd" means the ground separation, "KS" means the key-scan refine, and GT for application of ground truth pose}
            \label{tab:supp_abla_tab}

         \end{table}

         \begin{figure*}[t]
            \subfigure[Ground truth map]{
               \centering
               \includegraphics[width=0.48\textwidth]{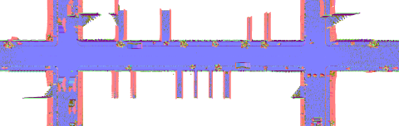}
            }
            \subfigure[Ours with GT pose]{
               \centering
               \includegraphics[width=0.48\textwidth]{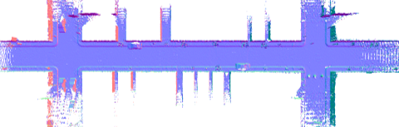}
             }
      
             \subfigure[Shine With GT pose]{
               \centering
               \includegraphics[width=0.5\textwidth]{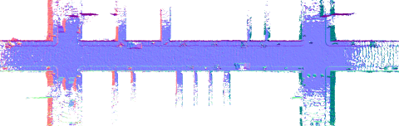}
            }
            \subfigure[Vdbfusion with GT pose]{
                \centering
                \includegraphics[width=0.5\textwidth]{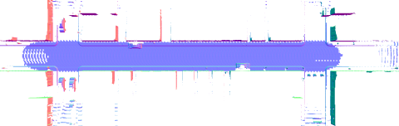}
             }
            \caption{The mapping result with ground truth pose on the MaiCity~\cite{vizzo2021icra} dataset. We present the bird-eye view of the results, indicating that our method can reconstruct a complete and smooth map.}
            \label{fig:supp_map_birdview}

         \end{figure*}

         \section{Odometry Evaluation on KITTI Dataset}
\begin{figure*}[t]
            \begin{center}
               \includegraphics[width=\textwidth]{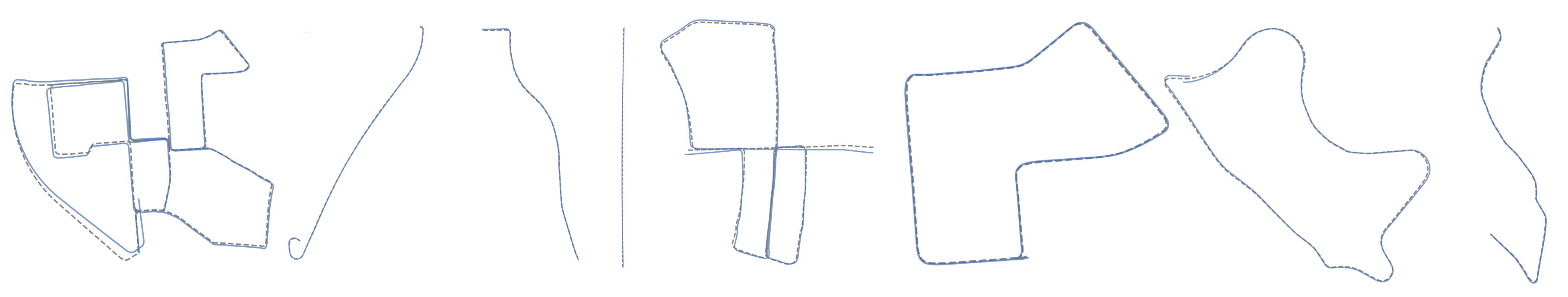}
            \end{center}
            \caption{The qualitative results of our odometry on KITTI~\cite{geiger2012cvpr} dataset. From left to right, we list the results of sequences 00, 01, 03, 04, 05, 07, 09, 10. The dashed line corresponds to the ground truth and the blue line to our odometry method.}
            \label{fig:supp_odo_kitti}

         \end{figure*}

         In this section, we present the odometry evaluation on KITTI~\cite{geiger2012cvpr} dataset. As can be seen from~\cref{tab:track_kitti}, our odometry results show comparative performance compared to the non-learning-based method and outperform them on some sequences. Compared to the learning-based method, our method does not need to be pre-trained by numerous labeled data, and it can be directly employed in other environments, where some existing learning-based methods fail. This is important when we lack adequate data and ground truth labels or explore unknown environments. We also present our qualitative results on~\cref{fig:supp_odo_kitti}. Our odometry process shows the ability of generalization on different sequences and large-scale environments.
      
         \section{Additional Ablation Study}
      
\begin{figure*}[t]
            \begin{center}
               \subfigure[Trajectory for MaiCity Dataset\cite{vizzo2021icra}]{
               \centering
               \includegraphics[width=0.475\textwidth]{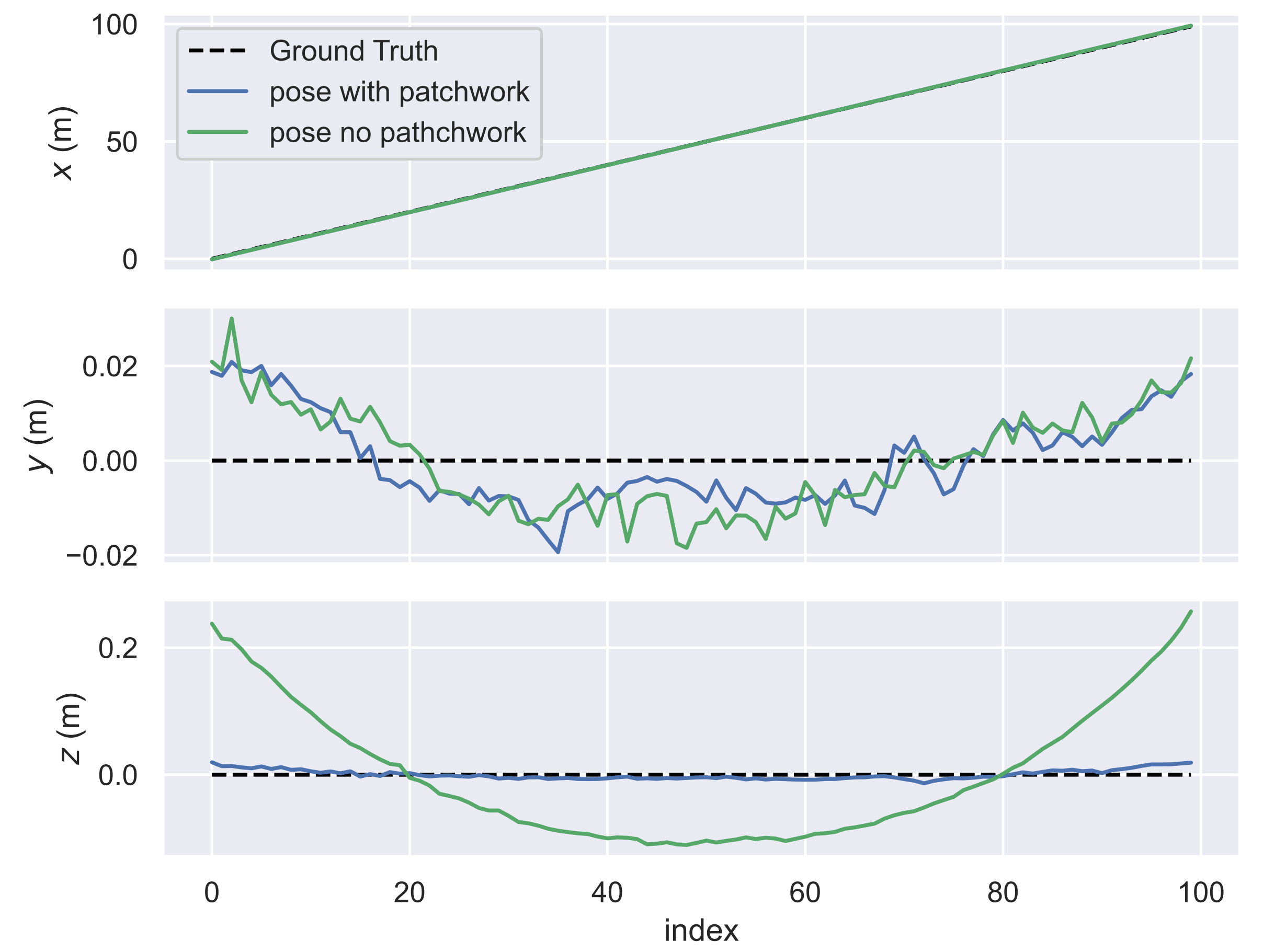}
            }
               \subfigure[Trajectory for Newer College Dataset\cite{ramezani2020iros}]{
               \centering
               \includegraphics[width=0.475\textwidth]{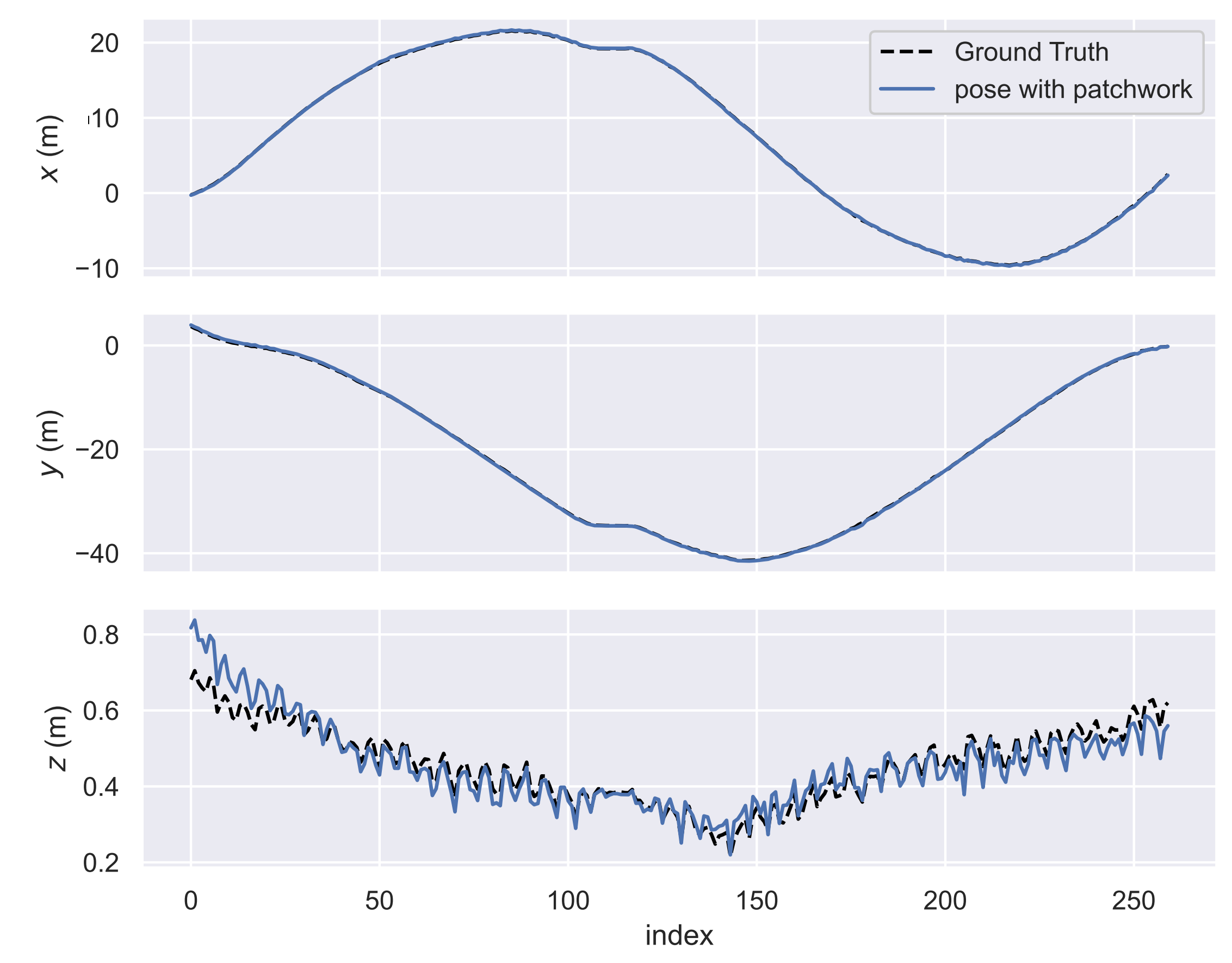}
            }
            \end{center}
            \caption{Ablation study for ground separation in terms of trajectory. The blue line is the trajectory with ground separation, and the green line is the one without ground separation. The dashed line represents the ground truth trajectory.}
            \label{fig:supp_abla_patch_traj}

         \end{figure*}

         We show the full table of ablation study on ~\cref{tab:supp_abla_tab} concerning the ground separation, key-scan refine, and application of ground truth pose. First, the ground separation can directly improve the odometry result, especially at the z-axis, where the ground separation takes effect. The qualitative results in ~\cref{fig:supp_abla_patch_traj} also prove its indispensable, and we can see that our method stays consistent at the z-axis on both datasets. Second, the key-scan refine can greatly improve the mapping quality when no ground truth pose is applied. Although this improvement becomes slight when the ground truth pose is applied, we still adopt this strategy as it can help us reconstruct a smooth and complete map. Third, the ground truth pose plays a significant role in mapping, especially for loops, which usually cause overlapping meshes. Dealing with loop detection is an important task for our future work.
      
      \begin{figure}[!t]
         \begin{center}

            \subfigure[Time vs Acc. on Maicity]{
               \centering
               \includegraphics[width=0.22\textwidth]{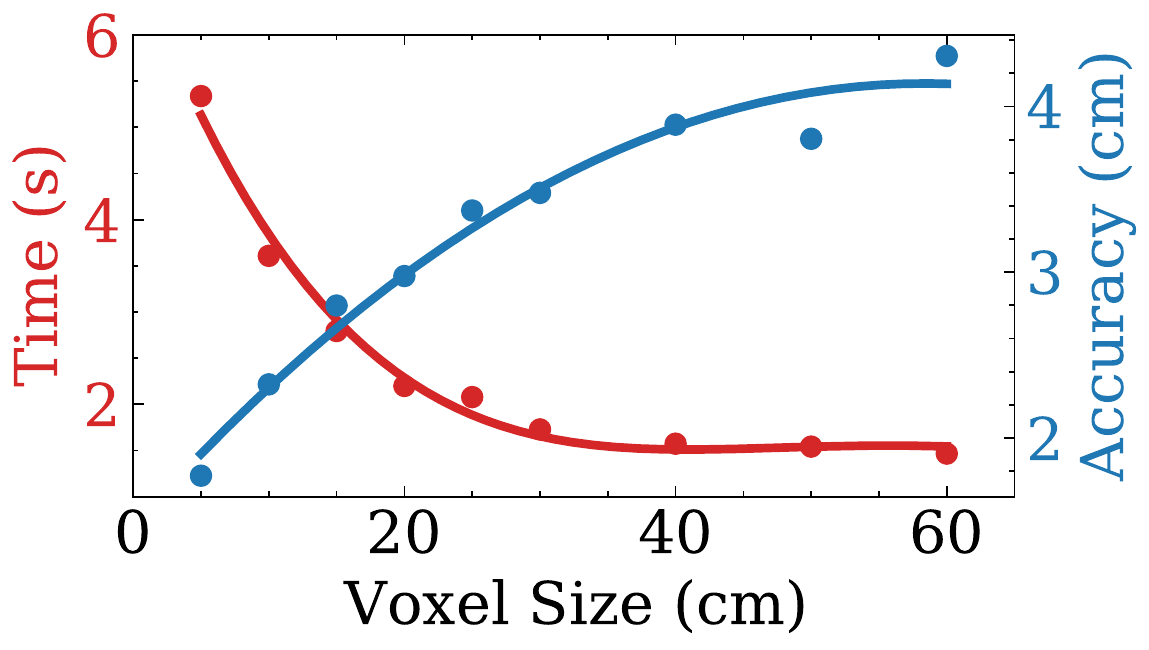}
            }
            \subfigure[Mem. vs Acc. on Maicity]{
               \centering
               \includegraphics[width=0.22\textwidth]{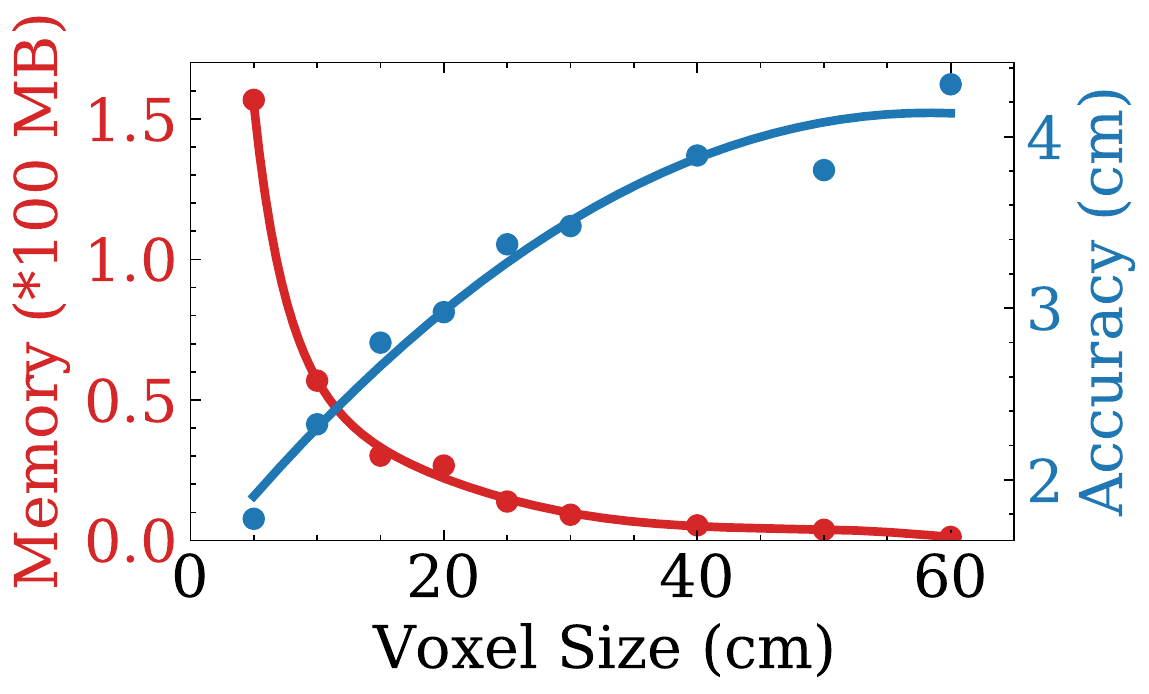}
            }
            
         \end{center}
         \caption{Study on voxel size v.s. processing time, memory consumption and accuracy distance on Maicity~\cite{vizzo2021icra}.}
         \label{fig:supp_ablavoxsize_mai}

      \end{figure}

      \begin{figure}[!t]
         \begin{center}
            \subfigure[Maicity]{
               \centering
               \includegraphics[width=0.22\textwidth]{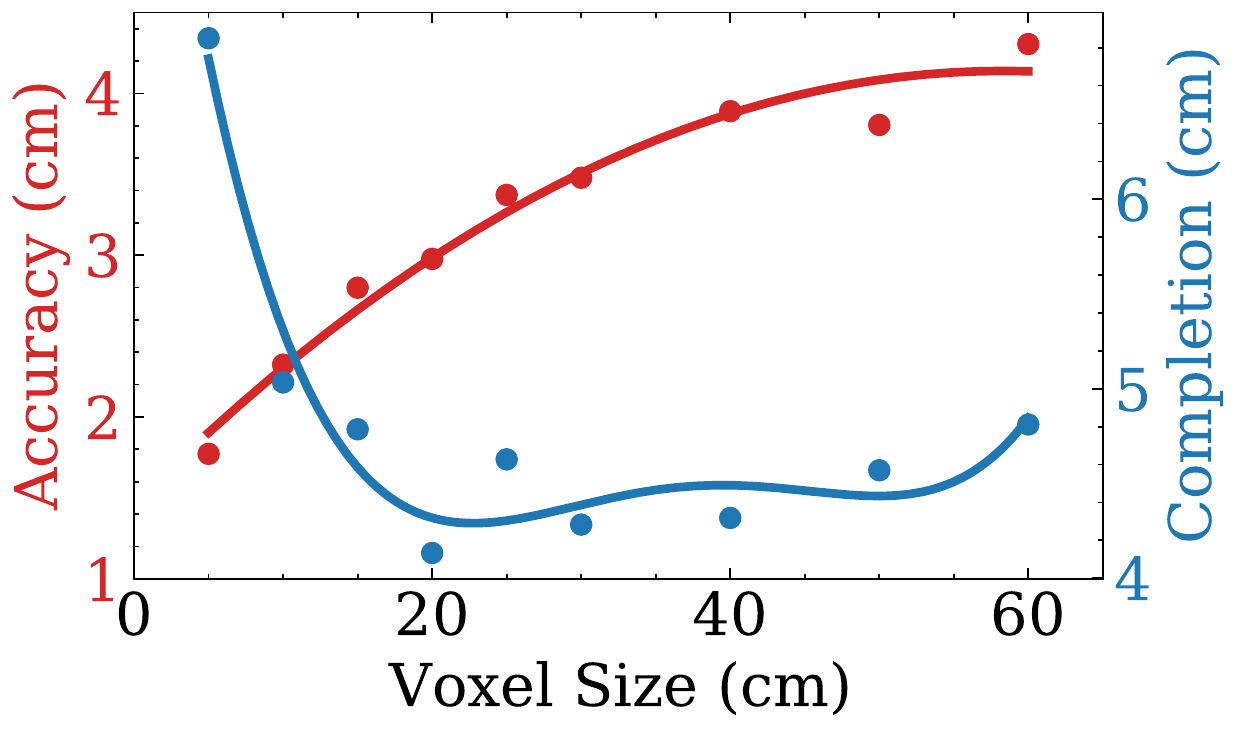}
            }
            \subfigure[Newer College]{
               \centering
               \includegraphics[width=0.22\textwidth]{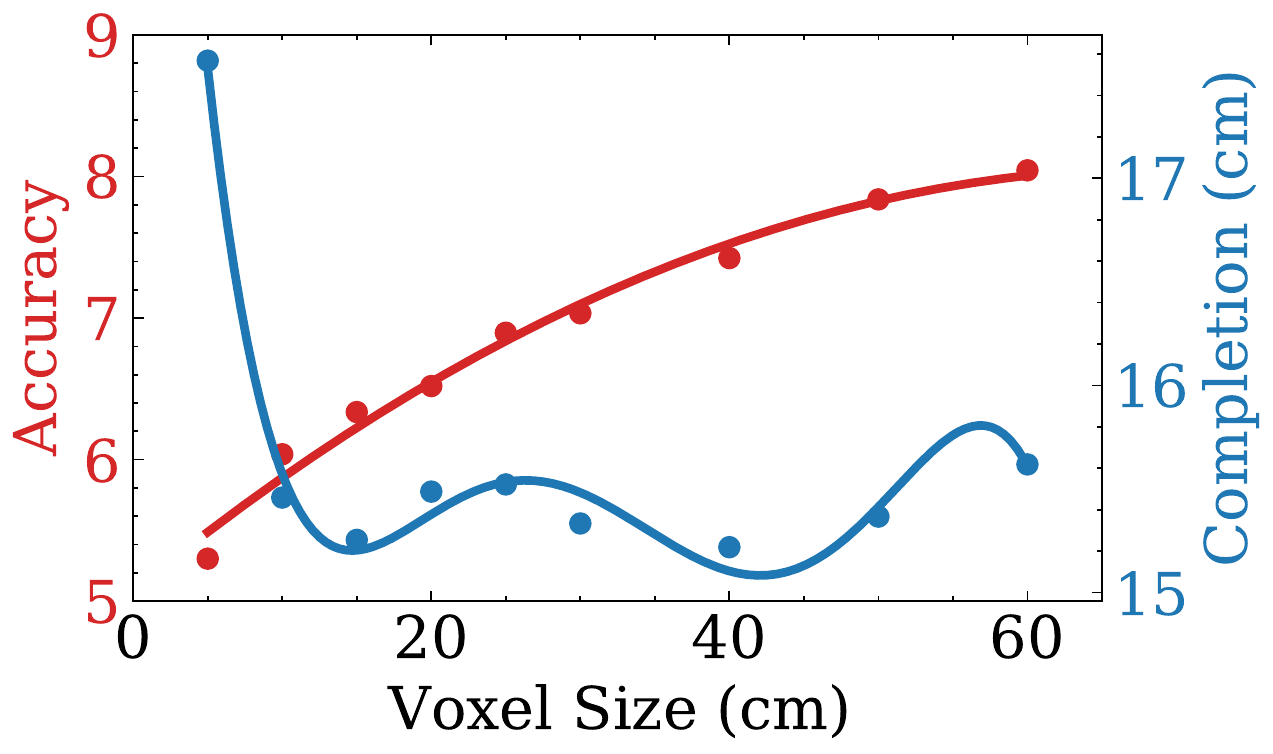}
            }
         \end{center}
         \caption{Study on voxel size v.s. accuracy distance and completion distance on Maicity~\cite{vizzo2021icra} and Newer College~\cite{ramezani2020iros}}
         \label{fig:supp_ablavoxsize2}

      \end{figure}

      \begin{figure}[!t]
         \begin{center}
   
            \subfigure[Trajectory Accuracy]{
               \centering
               \includegraphics[width=0.22\textwidth]{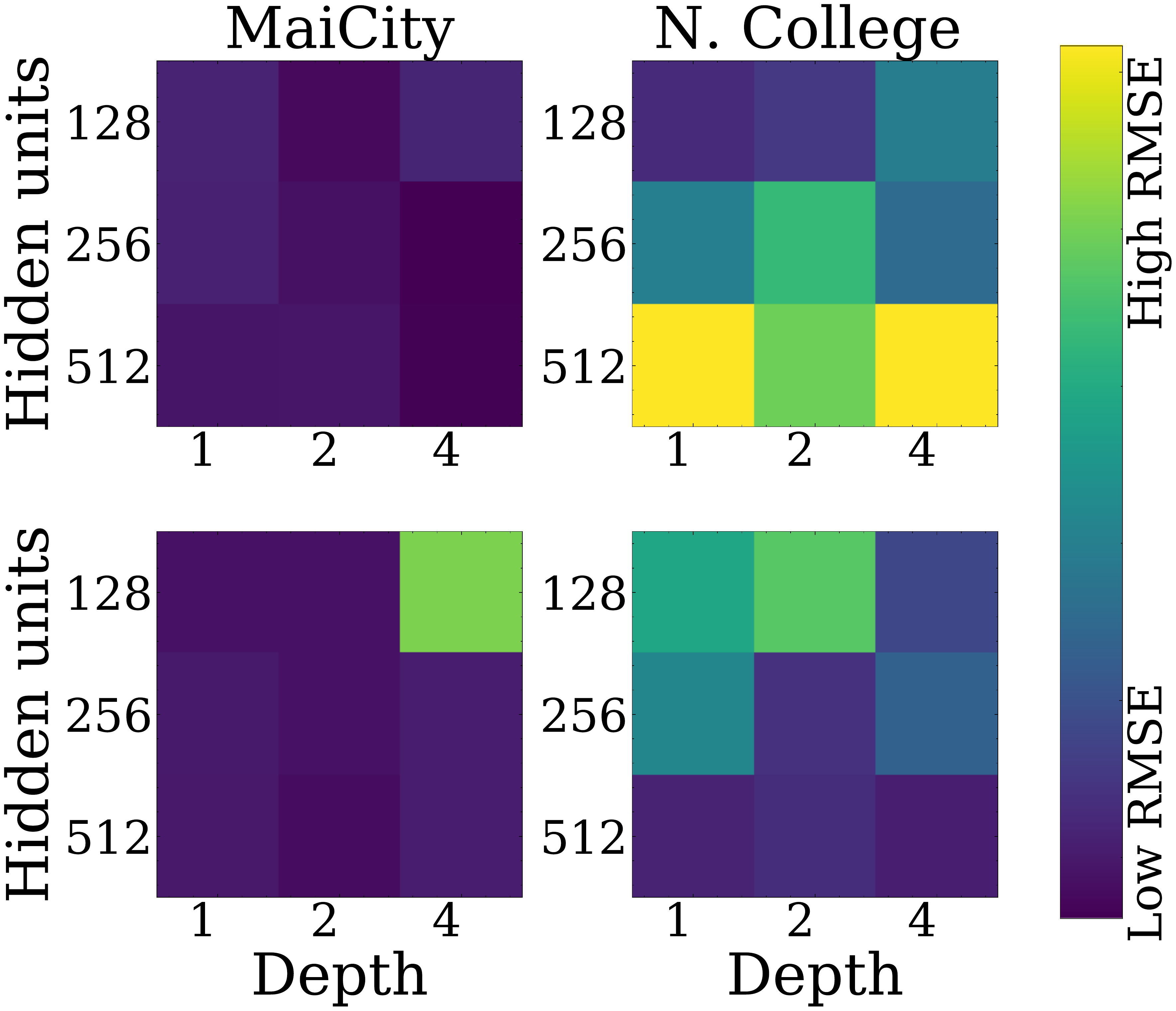}
            }
            \subfigure[Mapping Accuracy]{
               \centering
               \includegraphics[width=0.22\textwidth]{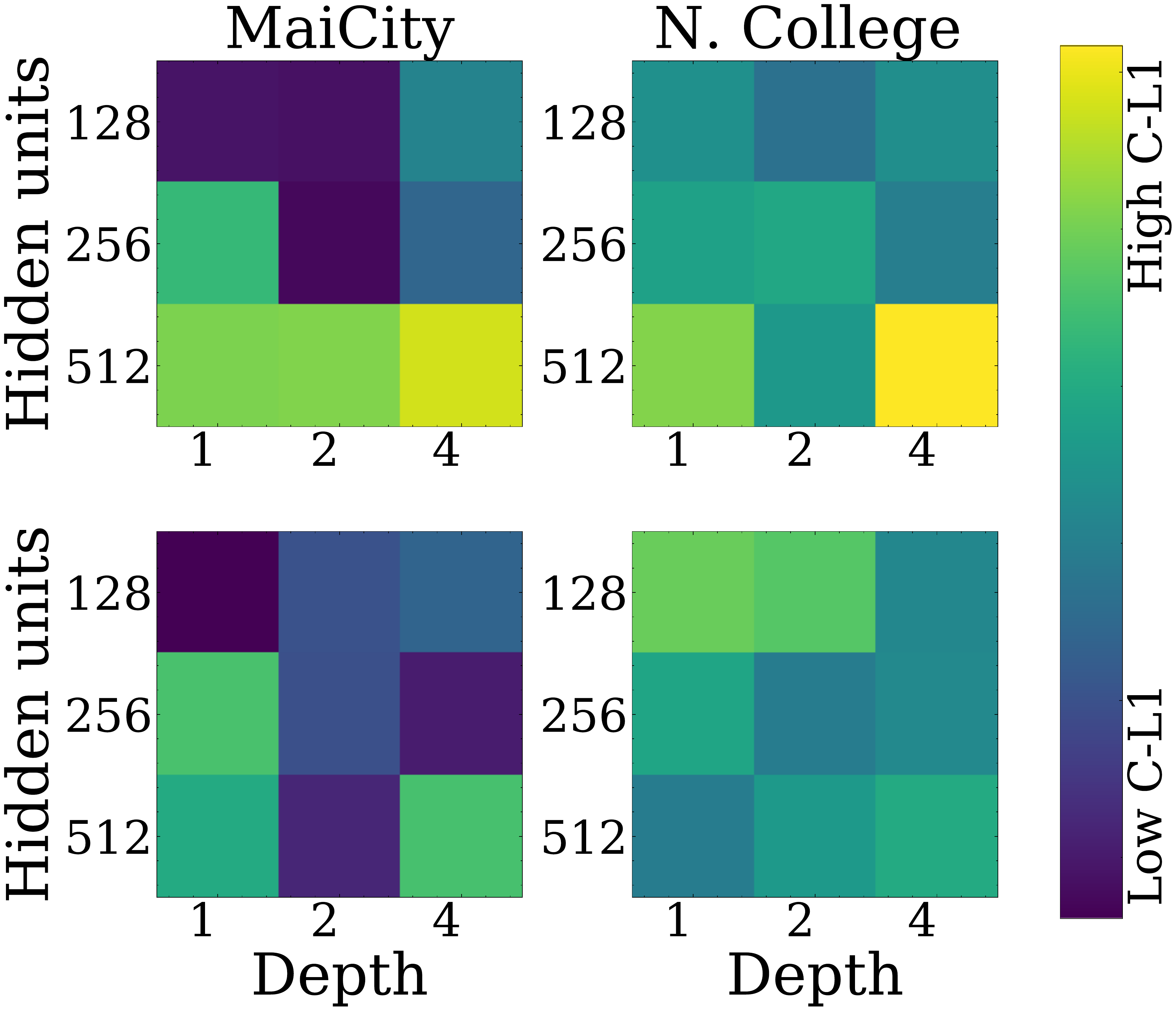}
            }
         \end{center}
         \caption{Ablation study for Network architecture and Embedding length. First low represent 8-embedding length, and second row is 16-length.}
         \label{fig:ablanetwork}

      \end{figure}

         We then complement in ~\cref{fig:supp_ablavoxsize_mai} the effect of voxel size on Maicity\cite{vizzo2021icra} dataset on the processing time, accuracy distance, and memory consumption. The two lines cross at voxel size between 15\,cm and 20\,cm. We choose 20\,cm as our choice for the reason that the processing time still decreases a lot while the accuracy remains. As indicated in ~\cref{fig:supp_ablavoxsize2}, we find that the Chamfer-L1 distance stays almost invariant, as the completion distance decrease with a larger voxel size. A smaller voxel size brings finer reconstruction while a larger voxel size can make it more complete. Similarly, we choose 20\,cm as the voxel size since the completion distance stays almost constant.
      
         We explore here the influence of network architecture (i.e., hidden units and depth) and embedding length. Figure \ref{fig:ablanetwork} show the normalized result of RMSE, Chamfer-L1 distance for various embedding length and network. During our study of the processing time, we found that a more profound and more hidden units network consumes more time while the embedding length affects little. We can also find from the figure that although a short embedding length can sometimes achieve exceptional results, it is unstable with the change of network. We choose 16 as our embedding length as it generalizes well and does not lower time efficiency. For the network, we use 2 layers deep and 256 hidden units architecture for it performs well in tracking as well as the mapping on two datasets while keeping time efficiency.

   \end{document}